\documentclass[runningheads]{llncs}

\usepackage[mobile]{eccv}

\usepackage{eccvabbrv}
\usepackage{makecell}

\usepackage{graphicx}
\usepackage{booktabs}

\usepackage[accsupp]{axessibility}  

\usepackage{hyperref}
\usepackage{color, colortbl}
\usepackage{stfloats}
\usepackage{enumitem}
\usepackage{tabularx}
\usepackage{xstring}
\usepackage{multirow}
\usepackage{xspace}
\usepackage{url}
\usepackage{subcaption}
\usepackage{xcolor}
\usepackage[hang,flushmargin]{footmisc}
\usepackage{algorithm}
\usepackage{algorithmic}
\usepackage{wrapfig}
\usepackage{placeins}

\usepackage{orcidlink}

\begin{document}

\title{GaMO: Geometry-aware Multi-view Diffusion Outpainting for Sparse-View 3D Reconstruction} 

\titlerunning{Abbreviated paper title}

\author{Yi-Chuan Huang \and Hao-Jen Chien \and Chin-Yang Lin \and Chih-Yu Chang \and Ying-Huan Chen \and Yu-Lun Liu}

\authorrunning{Y.-C.~Huang et al.}

\institute{National Yang Ming Chiao Tung University\\
\email{yichuanh.cs12@nycu.edu.tw, yulunliu@cs.nycu.edu.tw}}

\maketitle

\begin{center}
  \includegraphics[width=\textwidth]{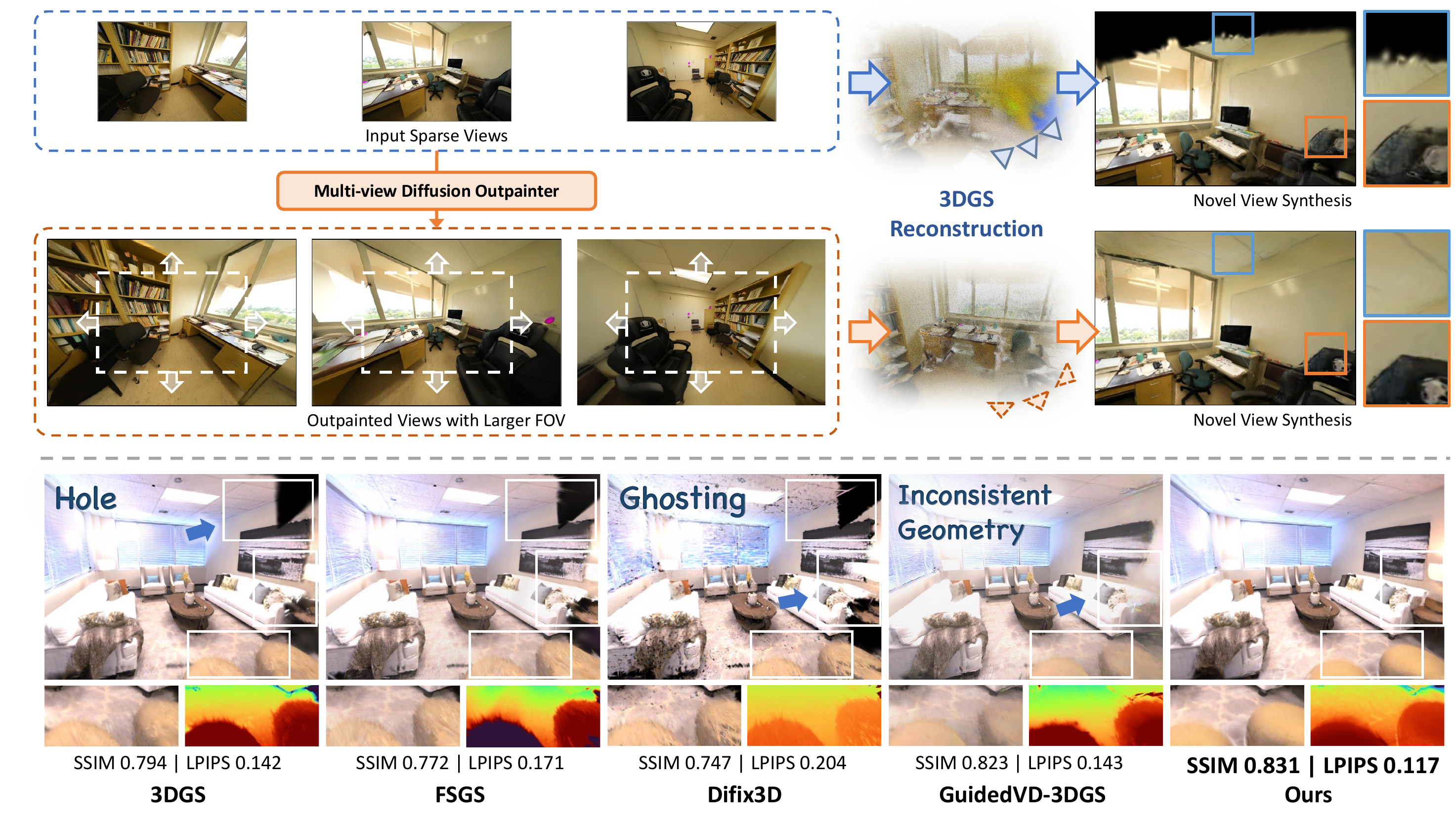}
  \captionsetup{type=figure}
  \caption{
\textbf{Overview and comparison.}
\emph{(Top)} Our method, \textbf{GaMO} (\textbf{G}eometry-\textbf{a}ware \textbf{M}ulti-view Diffusion \textbf{O}utpainter), expands sparse input views into wide-FOV outpainted views via a multi-view diffusion model, which are then used to refine 3D Gaussian Splatting (3DGS)~\cite{kerbl20233d} reconstruction, producing high-fidelity novel views with improved geometric consistency and visual clarity. 
\emph{(Bottom)} Qualitative comparison with existing methods, including 3DGS~\cite{kerbl20233d}, FSGS~\cite{zhu2024fsgs}, Difix3D~\cite{wu2025difix3d+}, and GuidedVD-3DGS~\cite{zhong2025taming}. Previous approaches suffer from \textit{holes}, \textit{ghosting}, or \textit{inconsistent geometry} when trained with sparse inputs. In contrast, our method effectively mitigates these artifacts and achieves superior image quality.
}
  \label{fig:teaser}
\end{center}

\begin{abstract}
Recent 3D reconstruction methods achieve impressive results with dense multi-view imagery but struggle when only a few views are available. Various approaches, including regularization techniques, semantic priors, and geometric constraints, have been implemented to address this challenge. Recent diffusion-based approaches further improve performance by generating novel views to augment training data. Despite this progress, we identify three critical limitations in current state-of-the-art approaches: (i) inadequate coverage beyond known view peripheries, (ii) geometric inconsistencies across generated views, and (iii) computational inefficiency due to expensive pipelines. We introduce \textbf{GaMO} (\textbf{G}eometry-\textbf{a}ware \textbf{M}ulti-view \textbf{O}utpainter), a framework that reformulates sparse-view reconstruction through multi-view outpainting. Instead of generating new viewpoints, GaMO expands the field of view from existing camera poses, which inherently preserves geometric consistency while providing broader scene coverage. Our approach employs multi-view conditioning and geometry-aware denoising strategies in a zero-shot manner without training. Extensive experiments on Replica, ScanNet++, and Mip-NeRF 360 demonstrate strong reconstruction performance across sparse-view settings (3, 6, and 9 input views). Notably, our method is significantly more efficient than existing diffusion-based approaches, reducing the overall runtime to within 10 minutes.
Project page: \url{https://yichuanh.github.io/GaMO/}
\end{abstract}

\section{Introduction}
\label{sec:intro}

Reconstructing complete 3D scenes from limited input views is a fundamental problem with numerous tangible applications, ranging from virtual property tours to immersive telepresence. However, it remains notoriously difficult, often resulting in broken geometry and visible visual artifacts. Previous approaches attempted to address the sparsity of input views through regularization, semantic priors, or geometric constraints~\cite{niemeyer2022regnerf,jain2021dietnerf,yang2023freenerf,zhu2024fsgs,wang2023sparsenerf,li2024dngaussian}. These methods remain limited in handling unobserved regions. 

Recently, diffusion-based approaches~\cite{zhong2025taming,wu2025genfusion,wu2024reconfusion,wu2025difix3d+,anciukevicius2023renderdiffusion} have generated novel views to improve the reconstruction quality for sparse observations. Nevertheless, these methods show three fundamental limitations: (1) novel view generation mainly focuses on enhancing angular coverage of existing geometry and often overlooks the extension beyond the periphery, leaving persistent \textbf{holes} and \textbf{ghostings} in the reconstruction; (2) geometric and photometric \textbf{inconsistencies} across novel and input views inevitably become prominent as view overlap increases due to internal diffusion variations; (3) novel view generation requires elaborate trajectory planning and camera pose sampling, making the process \textbf{time-consuming}.

\begin{figure}[t]
    \centering
    \includegraphics[width=1\columnwidth]{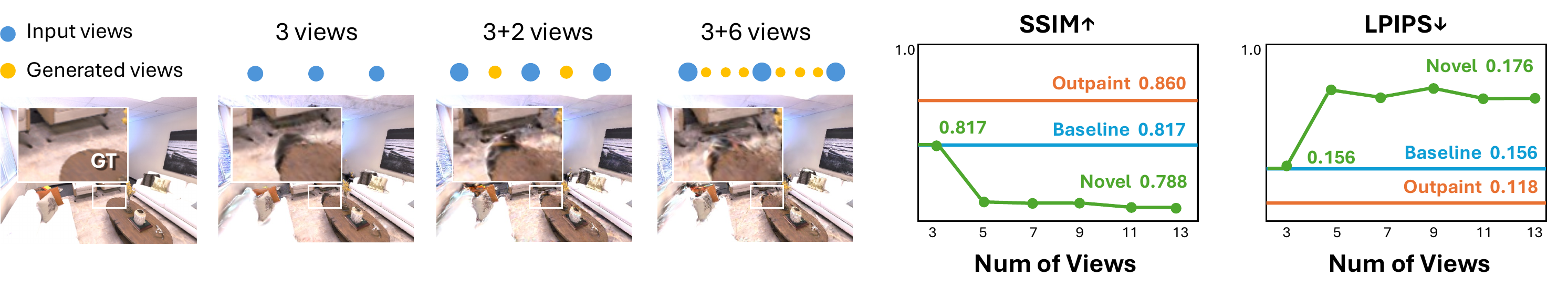}
\caption{
\textbf{Motivation: Outpainting vs. diffusion-generated novel views.}
Using multi-view diffusion~\cite{cao2025mvgenmaster}, we train 
3DGS~\cite{kerbl20233d} with three settings: interpolated views (\textcolor{green!70!black}{green}), a 3-view baseline (\textcolor{cyan}{blue}), and outpainting (\textcolor{orange}{orange}). 
\textbf{Top:} Adding more diffusion-generated views (3--13) may introduce geometric inconsistencies under sparse-view conditions. 
\textbf{Bottom:} SSIM/LPIPS show a similar trend: increasing generated views can degrade reconstruction quality, while outpainting remains more stable.
}
    \label{fig:moti} 
\end{figure}

Recent multi-view diffusion models incorporate geometry-aware priors and achieve strong geometric consistency across generated views. However, we find that directly using them to expand camera viewpoints is not suitable under extremely sparse-view settings. Fig.~\ref{fig:moti} shows an experiment where three input views are used to generate interpolated novel views via diffusion, which are then used to train 3DGS~\cite{kerbl20233d}. Increasing the number of generated views (from 3 to 13 total views) degrades reconstruction quality, leading to lower SSIM and higher LPIPS. This observation suggests that generating additional viewpoints is not always beneficial under sparse-view settings. 

Instead, we observe that \textit{outpainting}, rather than novel view generation, better leverages the geometry-aware priors of multi-view diffusion models and provides a more suitable paradigm for sparse-view 3D reconstruction. By extending content around existing input views, outpainting expands spatial coverage while preserving geometric consistency. Based on this insight, we propose \textbf{GaMO} (\textbf{G}eometry-\textbf{a}ware \textbf{M}ultiview \textbf{O}utpainting), which (1) expands the field of view (FOV) to cover unobserved regions while preserving geometric consistency, (2) avoids fusing multiple hallucinated views in 3D space, reducing misalignment artifacts, and (3) reconstructs scenes efficiently through a single outpainting pass, achieving tens-of-times speedup over video diffusion-based methods. We evaluate our approach on Replica~\cite{straub2019replica}, ScanNet++~\cite{yeshwanth2023scannet++}, and Mip-NeRF 360~\cite{barron2023mipnerf360}, demonstrating competitive performance with improved perceptual quality.

\noindent We summarize our contributions as follows:
\begin{itemize}
\item We establish outpainting as a superior paradigm for sparse-view reconstruction, eliminating common issues including holes, ghosting artifacts, and geometric inconsistencies.
\item We develop a geometry-aware outpainting approach with novel conditioning and denoising strategies in zero-shot manner without finetuning.
\item We achieve competitive performance across the Replica, ScanNet++, and Mip-NeRF 360 datasets. Our method provides strong geometric accuracy and perceptual quality across 3, 6, and 9 input views, while significantly reducing reconstruction time to under 10 minutes.

\end{itemize}

\section{Related Work}
\label{sec:related}

\paragraph{Sparse-view 3D Gaussian Splatting.}
While 3DGS~\cite{kerbl20233d} achieves remarkable quality with dense inputs, sparse-view reconstruction remains challenging, particularly for indoor scenes~\cite{li2024genrc}. Recent methods employ depth regularization~\cite{li2024dngaussian,chung2023depth}, proximity-guided unpooling~\cite{zhu2024fsgs}, dual-field co-regularization~\cite{zhang2024cor}, structural regularization via random Gaussian dropping~\cite{park2025dropgaussian}, matching-prior-based structure consistency~\cite{peng2024scgaussian}, and robust handling of unposed inputs~\cite{lin2025longsplat}. Diffusion-augmented approaches use generative models to synthesize pseudo-views for enhanced training supervision~\cite{kong2025generative,yang2024gaussianobject}. Feed-forward approaches leverage cost volumes~\cite{chen2024mvsplat,charatan2024pixelsplat}, Gaussian bundle adjustment~\cite{fan2024instantsplat}, depth-integrated splatting~\cite{xu2025depthsplat}, multi-view stereo features~\cite{liu2024mvsgaussian}, or transformer architectures~\cite{wang2024freesplat,xu2025freesplatter,tang2024hisplat,min2024epipolar,szymanowicz2024splatter}. Methods combining depth priors include DN-Splatter~\cite{turkulainen2025dn} with depth and normal cues, SplatFields~\cite{mihajlovic2024splatfields} regularizing spatial autocorrelation, and large model priors~\cite{yu2024lm,wang2024use,he2025see,shih2025prior}. While these methods \emph{regularize} 3D representations, our work \emph{augments} training data through geometry-aware outpainting for more complete scene coverage.

\paragraph{Multi-view diffusion models for 3D.}
Multi-view diffusion enables consistent 3D generation through multi-view attention~\cite{shi2023mvdream,wang2023imagedream}, synchronized volume attention~\cite{liu2023syncdreamer}, orthogonal view generation~\cite{shi2023zero123++}, and cross-domain diffusion~\cite{long2024wonder3d}. Recent advances enforce consistency via 3D feature unprojection~\cite{yang2024consistnet}, epipolar attention~\cite{huang2024epidiff,kuang2024collaborative}, depth-guided attention~\cite{hu2024mvd}, differentiable rasterization~\cite{lu2024direct2}, and pose-free dense generation~\cite{tang2024mvdiffusion++}. Large multi-view Gaussian models enable high-resolution content creation~\cite{tang2024lgm}. Video diffusion models provide temporal consistency for multi-view synthesis~\cite{voleti2024sv3d,gao2024cat3d,kwak2024vivid,zhang20244diffusion,chen2024mvsplat360,chen2024v3d,yu2024viewcrafter}, with view-integrated attention~\cite{xu2024cavia} and multi-view video generation~\cite{li2024vivid} further improving spatiotemporal coherence. Additional methods include mesh generation~\cite{xu2024instantmesh}, epipolar constraints~\cite{li2024era3d}, combined 2D-3D priors~\cite{liu2024one}, correspondence-aware attention~\cite{Tang2023mvdiffusion}, and 3D feature fields~\cite{chan2023generative}. These methods generate \emph{novel views} from different poses. Our work performs \emph{multi-view outpainting} to expand field-of-view of existing views, maintaining stronger geometric consistency for sparse-view scene reconstruction.

\paragraph{Diffusion priors for 3D reconstruction.}
Diffusion models provide learned priors through Score Distillation Sampling~\cite{poole2022dreamfusion,wang2023prolificdreamer,liang2024luciddreamer}. Improvements address over-smoothing~\cite{lukoianov2024score}, mode collapse~\cite{wang2024taming}, provide unified frameworks~\cite{mcallister2024rethinking}, classifier-based distillation~\cite{yu2023text}, ODE-based sampling~\cite{wu2024consistent3d,lee2024dreamflow}, and optimize both 3D models and priors~\cite{yang2024learn,chen2024vividdreamer}. Video diffusion serves as powerful priors~\cite{liu2024reconx,melas20243d,yi2024gaussiandreamer}. Reconstruction methods use multi-view conditioning~\cite{wu2024reconfusion}, pseudo-observation enhancement~\cite{liu2024deceptive}, scene-grounding guidance~\cite{zhong2025taming}, iterative refinement~\cite{wu2025genfusion,wu2025difix3d+,liu20243dgs}, inline prior-guided score matching under sparse views~\cite{wang2024use}, visibility-guided decompositional reconstruction~\cite{ni2025decompositional}, and various coupling strategies~\cite{xu2024bayesian,hui2024microdiffusion,muller2024multidiff,liu2024meshformer,xue2024human}. Native 3D diffusion includes latent approaches~\cite{lan2024ln3diff,he2024gvgen,wu2024direct3d,wu2024unique3d,yang2024scenecraft} and RL finetuning~\cite{xie2024carve3d}. While these methods generate additional views or provide guidance, they face multi-view \emph{inconsistency}. Our insight: diffusion models suit \emph{outpainting known views} better than hallucinating novel perspectives, maintaining stronger geometric grounding.

\paragraph{Geometry-aware generation.}
Geometric consistency leverages Pl\"ucker coordinates for camera conditioning~\cite{xu2023dmv3d,kant2024spad,zhang2024cameras,xu2024camco,ji2025campvg,he2024cameractrl,bahmani2025vd3d} and epipolar constraints or voxel representations~\cite{tu2023imgeonet} for multi-view consistency~\cite{huang2024epidiff,ye2024diffpano,wang2024mvdd,kupyn2025epipolar,muller2024multidiff,xu20243difftection,zheng2024cami2v}. Camera motion control in video diffusion extends these representations to temporal settings~\cite{wang2024motionctrl}. Joint pixel-level image and depth synthesis further improves geometric fidelity~\cite{guizilini2025zero}. Depth and normal conditioning proves critical~\cite{lu2024direct2,long2024wonder3d,ke2024repurposing,fu2024geowizard,patni2024ecodepth,hu2024mvd,duan2024diffusiondepth,wang2024depthanywhere,qiu2024richdreamer}. Recent panoramic generation~\cite{zhang2024continuous,wang2024360dvd} and video outpainting~\cite{yu2025unboxed} typically operate in 2D or single-view scenarios. Our approach uniquely combines \emph{multi-view outpainting} with \emph{geometry awareness} through coarse 3DGS rendering, opacity-based masking, and noise resampling for consistent, geometrically plausible FOV expansion.

\paragraph{Outpainting and FOV expansion.}
Diffusion-based outpainting includes panoramic methods~\cite{zhang2024continuous,wu2023panodiffusion,kalischek2025cubediff,yuan2025camfreediff,feng2023diffusion360,shi2023fishdreamer,yu2024shadow,zhang2024taming,lu2024autoregressive,liu2024panofree} and restoration tasks~\cite{tsai2025lightsout,liu2025corrfill}. Video outpainting methods leverage input-specific adaptation~\cite{wang2024your} and temporal diffusion~\cite{zhang2024avid}. For 3D scenarios, methods employ visibility-aware inpainting~\cite{liu2024novel,wu2025aurafusion360}, generative scene completion via grid priors~\cite{weber2024nerfiller} or interactive extrapolation~\cite{yu2025wonderworld}, reference-adapted diffusion for 3D inpainting~\cite{mirzaei2024reffusion}, video diffusion priors~\cite{liu2024novel}, NeRF-guided training~\cite{yu2024nerf}, iterative 3DGS updates~\cite{yu2025unboxed}, and multi-view SDS~\cite{chen2024mvip}. General sparse-view baselines include feed-forward prediction~\cite{charatan2024pixelsplat,chen2024mvsplat}, regularized optimization~\cite{paliwal2024coherentgs,xu2024sparp}, and NeRF-based methods~\cite{wang2023sparsenerf,yang2023freenerf,niemeyer2022regnerf,lin2025frugalnerf,su2024boostmvsnerfs}. These works require per-scene fine-tuning or focus on single-view outpainting. Our method performs \emph{zero-shot multi-view outpainting} using pre-trained MVGenMaster~\cite{cao2025mvgenmaster} with geometry-aware mechanisms ensuring cross-view consistency without scene-specific training.

\section{Preliminaries}

\paragraph{3D Gaussian Splatting }
(3DGS)~\cite{kerbl20233d} uses a collection of anisotropic 3D Gaussian primitives to present a scene. Each Gaussian is defined by its center position $\boldsymbol{\mu} \in \mathbb{R}^3$, a 3D covariance matrix $\boldsymbol{\Sigma}$, an opacity value $\alpha \in [0,1]$, and spherical harmonic coefficients for view-dependent color. The covariance matrix is decomposed into a scaling vector $\mathbf{s} \in \mathbb{R}^3$ and rotation quaternion $\mathbf{q} \in \mathbb{R}^4$ as $\boldsymbol{\Sigma} = \mathbf{R}\mathbf{S}\mathbf{S}^T\mathbf{R}^T$, where $\mathbf{R}$ is derived from $\mathbf{q}$ and $\mathbf{S} = \text{diag}(\mathbf{s})$. The Gaussian function is:
\begin{equation}
\mathcal{G}(\mathbf{x}) = \exp\left(-\frac{1}{2}(\mathbf{x} - \boldsymbol{\mu})^T\boldsymbol{\Sigma}^{-1}(\mathbf{x} - \boldsymbol{\mu})\right).
\end{equation}

To render a given viewpoint, 3DGS projects each 3D Gaussian onto the 2D image plane, obtaining a 2D Gaussian $\mathcal{G}'(\mathbf{u})$, where $\mathbf{u}$ is pixel coordinates. The color of pixel $\mathbf{u}$ is computed via $\alpha$-blending of ordered Gaussians:
\begin{equation}
\mathbf{C}(\mathbf{u}) = \sum_{i \in \mathcal{N}} \mathbf{c}_i \sigma_i \prod_{j=1}^{i-1} (1 - \sigma_j),
\end{equation}
where $\mathcal{N}$ denotes the set of Gaussians overlapping pixel $\mathbf{u}$, sorted in depth order, $\mathbf{c}_i$ represents the color of the $i$-th Gaussian, and $\sigma_i = \alpha_i \mathcal{G}'_i(\mathbf{u})$ is the opacity contribution.

\paragraph{Diffusion Models}
generate samples through a learned denoising process that reverses a forward noising process. The forward process gradually adds Gaussian noise to data $\mathbf{x}_0$ over $T$ timesteps: $\mathbf{x}_t = \sqrt{\bar{\alpha}_t}\mathbf{x}_0 + \sqrt{1-\bar{\alpha}_t}\boldsymbol{\epsilon}$, where $\boldsymbol{\epsilon} \sim \mathcal{N}(\mathbf{0}, \mathbf{I})$ and $\bar{\alpha}_t$ is a predefined noise schedule. The reverse process learns to denoise $\mathbf{x}_t$ back to $\mathbf{x}_0$ by training a neural network $\epsilon_\theta$ to predict the noise at each timestep. The training objective is the simplified loss function:
\begin{equation}
\mathcal{L}_{\text{simple}} = \mathbb{E}_{t,\mathbf{x}_0,\boldsymbol{\epsilon}}\left[\|\boldsymbol{\epsilon} - \epsilon_\theta(\mathbf{x}_t, t, \mathbf{c})\|^2\right],
\end{equation}
where $\mathbf{c}$ represents conditioning information, and the model learns to minimize the mean squared error between the true noise $\boldsymbol{\epsilon}$ and the predicted noise. During inference, samples are generated by iteratively denoising from pure noise $\mathbf{x}_T \sim \mathcal{N}(\mathbf{0}, \mathbf{I})$ using the DDIM~\cite{song2021denoising} sampling process.

\section{Method}
\label{sec:method}

To address challenges in sparse-view 3D reconstruction, we perform geometry-aware outpainting by leveraging multi-view diffusion models.
By expanding the field-of-view (FOV) of input images, our method simultaneously fills holes, fixes blurred boundaries, and preserves geometric consistency without modifying existing content, resulting in a significantly simpler and faster reconstruction process.

\begin{figure}[t]
    \centering
    \includegraphics[width=1\columnwidth]{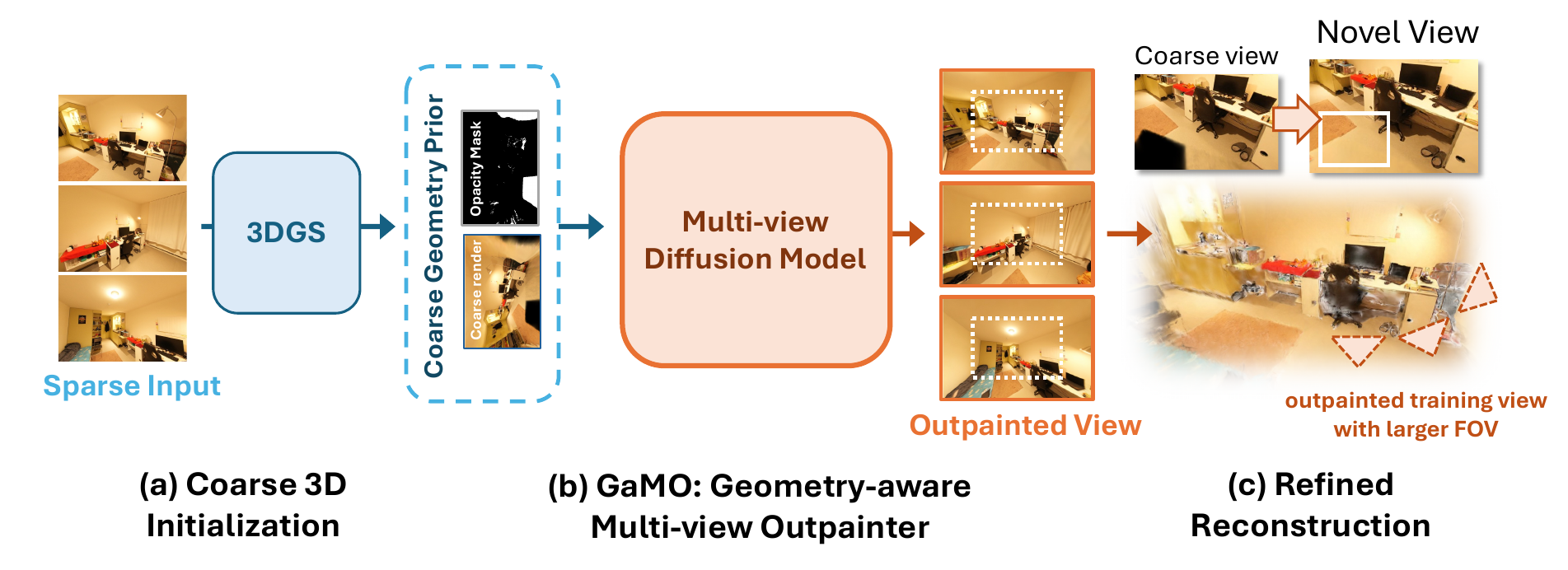}
    \caption{
    \textbf{Overview of Our Pipeline.}
    Given sparse input views, our method follows a three-stage process. 
    \textbf{(a) Coarse 3D Initialization:} We obtain geometry priors from initial 3D reconstruction, including an opacity mask and coarse render that provide essential structural cues. 
    \textbf{(b) Geometry-aware Multi-view Outpainter:} Using the geometry priors, GaMO generates outpainted views with enlarged FOV via a multi-view diffusion model. 
    \textbf{(c) Refined Reconstruction:} The outpainted views are used to refine the 3D reconstruction, resulting in improved completeness and consistency.
    }
     \label{fig:pipeline}
\end{figure}


As illustrated in Fig.~\ref{fig:pipeline}, our pipeline consists of three stages: coarse 3D initialization to obtain geometry priors (Sec.~\ref{sec:coarse_init}), geometry-aware multi-view outpainting to generate enlarged FOV views (Sec.~\ref{sec:GaMO}), and refined 3D reconstruction using the outpainted views (Sec.~\ref{sec:Refine}).

\FloatBarrier
\begin{figure}[t]
    \centering
    \includegraphics[width=\columnwidth]{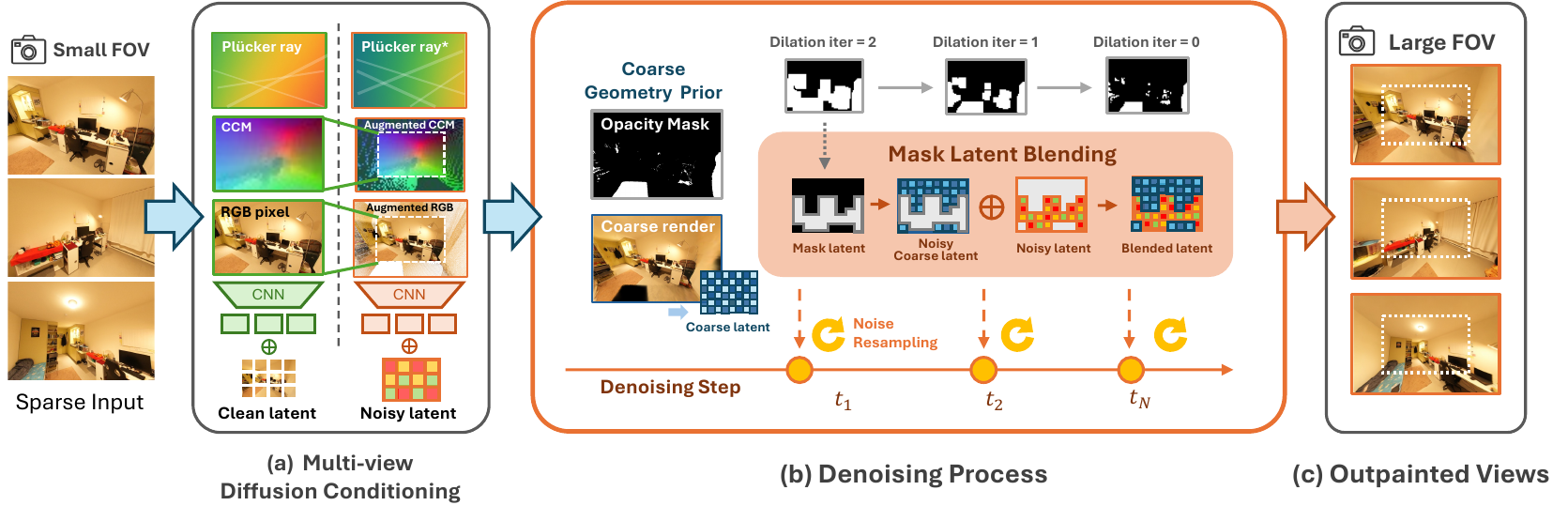}
    \caption{\textbf{Overview of GaMO (Geometry-aware Multi-view Diffusion Outpainter).}
    (a) Multi-view Diffusion Conditioning: Sparse input views are encoded into clean latents and combined with multi-view conditions, including Plücker ray embeddings for input views ($\mathcal{P}_r$) and the target view with enlarged FOV ($\mathcal{P}_t^*$), along with original and augmented Canonical Coordinate Map (CCM) and RGB, to provide both geometric and appearance cues for diffusion model conditioning.
    (b) Denoising Process: Coarse geometry priors (opacity mask and coarse render) guide the denoising through mask latent blending performed at multiple timesteps ($t_1, t_2, ..., t_N$) with progressive dilation and noise resampling, generating outpainted views with enlarged FOV (c).}
    \label{fig:GaMO}
\end{figure}

\subsection{Coarse 3D Initialization}
\label{sec:coarse_init}

To ensure geometric consistency in the diffusion model, we use DUSt3R~\cite{wang2024dust3r} to generate an initial point cloud and train a coarse 3DGS model to capture the scene geometry. Using this coarse model, we identify outpainting regions by rendering an opacity mask with a FOV wider than the input views. We also render a coarse color image to provide appearance priors for the outpainting process in Sec.~\ref{sec:GaMO}.

\paragraph{Opacity Mask.}
We enlarge the FOV by reducing the focal lengths with a scaling ratio $S_k < 1$ (i.e., $f_x' = f_x \times S_k$, $f_y' = f_y \times S_k$). For each target outpainted view, we first render an opacity map $\mathcal{O}$ by $\alpha$-blending the opacity values of the Gaussians:
\begin{equation}
\mathcal{O}(\mathbf{u}) = \sum_{i \in \mathcal{N}} \sigma_i \prod_{j=1}^{i-1} (1 - \sigma_j),
\end{equation}
where $\sigma_i = \alpha_i \mathcal{G}'_i(\mathbf{u})$ denotes the opacity contribution of the $i$-th Gaussian at pixel $\mathbf{u}$. The opacity mask $\mathcal{M}$ is then obtained by thresholding the opacity map with $\mathcal{M} = \mathbb{I}(\mathcal{O} < \eta_{\text{mask}})$, where $\eta_{\text{mask}}$ is a threshold value and $\mathbb{I}(\cdot)$ is the indicator function. Regions where $\mathcal{M} = 1$ correspond to areas with low opacity that require outpainting.

\paragraph{Coarse Rendering.}
We render a color image $I_{\text{coarse}}$ with the enlarged FOV from the coarse 3DGS model. This coarse rendering serves as a reference that provides geometric and appearance priors to the diffusion model, maintaining consistency between outpainted and existing scene content.

\subsection{GaMO: Geometry-aware Multi-view Diffusion Outpainter}
\label{sec:GaMO}

\noindent Our geometry-aware outpainting method operates through three key components: (1) multi-view conditioning that provides structural and appearance guidance; (2) mask latent blending that integrates coarse geometry priors during denoising; and (3) iterative mask scheduling with noise resampling that ensure smooth transitions. The model operates in latent space using DDIM sampling~\cite{song2021denoising} for efficient denoising.

\paragraph{Multi-View Conditioning.}
Given a set of sparse input RGB images $\{I_i\}_{i=1}^N$ and their corresponding camera parameters $\{\Pi_i\}_{i=1}^N$, our model generates outpainted views conditioned on camera representations, geometric correspondences, and appearance features, as illustrated in Fig.~\ref{fig:GaMO}(a).

For camera representation, we employ Plücker ray embeddings~\cite{xu2024dmv3d} that provide dense 6D ray parameterizations for each pixel, compactly encoding both ray origin and direction for geometry-aware reasoning. The embedding of each input view $\mathcal{P}_r$ is derived from its corresponding camera parameters $\Pi_r$, while the embedding of the outpainted view $\mathcal{P}_t^*$ uses the same camera parameters with scaled focal lengths $(f_x', f_y')$ to align with the enlarged FOV.

For geometric correspondence, we warp input RGB images and Canonical Coordinate 
Maps (CCM)~\cite{li2024sweetdreamer} to align with the expanded FOV by unprojecting pixels to 3D and 
reprojecting onto the outpainted camera plane, producing $\mathcal{C}_{r \rightarrow t}^{\text{warp}}$ 
and $I_{r \rightarrow t}^{\text{warp}}$. We then downsample the original inputs 
by factor $S_k$ and place them at the center of the warped features, creating 
augmented signals $I_{r \rightarrow t}^{\text{aug}}$ and $\mathcal{C}_{r \rightarrow t}^{\text{aug}}$ 
where the center preserves exact input information while the periphery retains 
warped geometric structure to guide outpainting.

For appearance features, the input RGB images are encoded through a variational autoencoder (VAE) to obtain clean latent features $\mathbf{z}_r$. The noisy latent features $\mathbf{z}_t$ are randomly generated and will be denoised to generate the outpainted views. 

All conditioning signals are processed through lightweight convolutional encoders. 
For input views, Plücker ray embeddings $\mathcal{P}_r$, CCM $\mathcal{C}_r$, 
and RGB images $I_r$ are jointly added to the clean latent features $\mathbf{z}_r$. 
For the target outpainted view, $\mathcal{P}_t^*$, $\mathcal{C}_{r \rightarrow t}^{\text{aug}}$, 
and $I_{r \rightarrow t}^{\text{aug}}$ are jointly added to the noisy latent 
features $\mathbf{z}_t$. We then condition the pre-trained diffusion model with 
the fused features to generate outpainted view latents in a zero-shot manner:
\begin{equation}
p_\theta(\mathbf{z}_t | \mathbf{z}_r, \mathcal{P}_r, \mathcal{C}_r, I_r, \mathcal{P}_t^*, \mathcal{C}_{r \rightarrow t}^{\text{aug}}, I_{r \rightarrow t}^{\text{aug}}),
\label{eq:diffusion_conditioning}
\end{equation}
where $\theta$ denotes the pre-trained model~\cite{cao2025mvgenmaster} parameters. These multi-view conditions ensure that the diffusion process maintains geometric consistency across views, even under an enlarged FOV.

\paragraph{Denoising Process with Mask Latent Blending.}
As the central component of our geometry-aware framework, mask latent blending integrates coarse geometry priors from the coarse 3D initialization (Sec.~\ref{sec:coarse_init}) into the diffusion loop. As outlined in Alg.~\ref{alg:gamo}, this process ensures that outpainted content respects existing scene structures while generating plausible peripheral regions. Fig.~\ref{fig:GaMO}(b) shows that the opacity mask $\mathcal{M}$ and coarse rendering $I_{\text{coarse}}$ provide consistent structural guidance throughout denoising.

At selected denoising timesteps $\{t_1, t_2, ..., t_N\}$, we perform \textit{mask latent blending} between the denoised latent and the coarse geometry prior. To ensure both latents share the same noise level, we add noise to the coarse latent, which is obtained by encoding the coarse rendering into latent space, before blending them using a latent-space mask $\mathcal{M}_{\text{latent}}$. The mask evolution is controlled by iterative mask scheduling (Sec.~\ref{sec:IMS}):
\begin{equation}
\mathbf{z}^{\text{blend}}_{t_k} = (1 - \mathcal{M}_{\text{latent}}^{(k)} ) \odot 
\mathbf{z}^{\text{coarse}}_{t_k} + \mathcal{M}_{\text{latent}}^{(k)} \odot \mathbf{z}_{t_k},
\label{eq:IMS}
\end{equation}
where $\mathbf{z}_{t_k}$ is the denoised latent, $\mathbf{z}^{\text{coarse}}_{t_k}$ 
is the coarse latent with matching noise level (Alg.~\ref{alg:gamo}, line 11), 
$\mathcal{M}_{\text{latent}}^{(k)}$ is the dilated mask at iteration $k$, and 
$\odot$ denotes element-wise multiplication.

\begin{wrapfigure}{R}{0.55\textwidth}
    \begin{minipage}{\linewidth}
    \vspace{-40pt} 
    \begin{algorithm}[H]
        \caption{Geometry-aware Multi-view Outpainter}
        \label{alg:gamo}
        \scriptsize
        \begin{algorithmic}[1]
            \STATE \textbf{Input:} Coarse render $I_{\text{coarse}}$, opacity mask $\mathcal{M}$
            \STATE \textbf{Output:} Outpainted views $\{S_j^{\text{out}}\}_{j=1}^M$
            \STATE \textbf{Setup:} Noise schedule $\Sigma = \{\sigma_1, \ldots, \sigma_T\}$
            \STATE \textbf{Setup:} latent blending iterations $\{t_1, \ldots, t_N\}$
            \STATE \textbf{Setup:} resampling iterations $R$
            \STATE $\mathbf{z}^{\text{coarse}} \gets \text{Encode}(I_{\text{coarse}})$
            \STATE $\mathbf{z}_T \sim \mathcal{N}(\mathbf{0}, \mathbf{I})$
            \FOR{$s = T, \ldots, 1$}
                \STATE $\mathbf{z}_{s-1} = \text{Denoise}(\mathbf{z}_s, \text{conditions})$ \hfill $\triangleright$ Eq.~\eqref{eq:diffusion_conditioning}
                \IF{$s \in \{t_1, \ldots, t_N\}$}
                    \STATE $\mathbf{z}_{s-1}^{\text{coarse}} = \text{AddNoise}(\mathbf{z}^{\text{coarse}}, \sigma_{s-1})$
                    \STATE $\mathbf{z}_{s-1}^{\text{blend}} = \text{IMS}(\mathcal{M}, \mathbf{z}_{s-1}, \mathbf{z}_{s-1}^{\text{coarse}})$ \hfill $\triangleright$ Eq.~\eqref{eq:IMS}
                    \STATE $\mathbf{z}_{s-1} = \mathbf{z}_{s-1}^{\text{blend}}$
                    \FOR{$r$ in $R$}
                        \STATE $\hat{\mathbf{z}}_0 = \text{Predict}(\mathbf{z}_{s-1})$
                        \STATE $\mathbf{z}_s^{\text{resamp}} \gets \text{AddNoise}(\hat{\mathbf{z}}_0, \sigma_s)$ \hfill $\triangleright$ Eq.~\eqref{eq:resamp}
                        \STATE $\mathbf{z}_{s-1} \gets \text{Denoise}(\mathbf{z}_s^{\text{resamp}}, \text{conditions})$
                    \ENDFOR
                \ENDIF
            \ENDFOR
            \STATE $\{S_j^{\text{out}}\}_{j=1}^M = \text{Decode}(\mathbf{z}_0)$
        \end{algorithmic}
    \end{algorithm}
    \vspace{-40pt}
    \end{minipage}
\end{wrapfigure}

\paragraph{Iterative Mask Scheduling and Noise Resampling.}
\label{sec:IMS}
To gradually integrate generated content with the existing geometric structure, Iterative Mask Scheduling progressively adjusts $\mathcal{M}_{\text{latent}}^{(k)}$ over iterations $k$ to control the ratio between outpainting and known coarse regions. The mask dilation is progressively reduced as denoising proceeds, allowing the model to first explore peripheral content and later refine geometry within coarse regions.

To maintain smooth transitions across blended regions, we perform noise resampling after each blending operation. After blending, we perform noise resampling $R$ times on the blended latent to eliminate boundary artifacts and ensure smooth integration between the coarse geometry and generated content (Alg.~\ref{alg:gamo}, lines 14–17). Specifically, we first predict the clean latent $\hat{\mathbf{z}}_0$ from the blended latent, then add noise back to the current timestep $t_k$:
\begin{equation}
\mathbf{z}^{\text{resamp}}_{t_k} = \sqrt{\bar{\alpha}_{t_k}} \hat{\mathbf{z}}_0 + \sqrt{1 - \bar{\alpha}_{t_k}} \boldsymbol{\epsilon},
\label{eq:resamp}
\end{equation}
where $\hat{\mathbf{z}}_0$ is the predicted clean latent from $\mathbf{z}^{\text{blend}}_{t_k}$ and $\boldsymbol{\epsilon} \sim \mathcal{N}(\mathbf{0}, \mathbf{I})$ denotes sampled Gaussian noise. This resampling prevents boundary artifacts and ensures smooth blending.

This framework ensures that outpainted regions seamlessly blend with known content while maintaining geometric plausibility, with the coarse 3DGS geometry providing structural guidance throughout the generation process. Importantly, it requires only inference without fine-tuning the backbone diffusion model.

\subsection{3DGS Refinement with Outpainted Views} \label{sec:Refine}
Given the original input views $\{I_i^{gt}\}_{i=1}^N$ and the generated outpainted views $\{S_j^{\text{out}}\}_{j=1}^M$ from Sec.~\ref{sec:GaMO}, we refine the 3DGS model by jointly optimizing with both sets of views. During training, we sample either an input view or an outpainted view for supervision at each iteration. 

\noindent\textbf{Loss for Input Views.}
We employ the standard 3DGS reconstruction loss~\cite{kerbl20233d} to ensure accurate reconstruction of the observed regions:
\begin{equation}
\mathcal{L}_{\text{input}} = (1-\lambda_s)\mathcal{L}_1(I_i, I^{gt}_i) + \lambda_s\mathcal{L}_{\text{D-SSIM}}(I_i, I^{gt}_i),
\label{eq:loss_input}
\end{equation}
where $I_i$ denotes the rendered image from input viewpoint, $I^{gt}_i$ is the ground truth input view, and $\lambda_s$ is a weighting factor that balances the $\mathcal{L}_1$ loss and structural similarity loss.

\noindent\textbf{Loss for Outpainted Views.}
Relying solely on reconstruction loss fails to fill unobserved regions and causes artifacts. We incorporate perceptual loss~\cite{johnson2016perceptual} $\mathcal{L}_{\text{LPIPS}}$ to provide balanced gradients across outpainted and original regions, effectively guiding training while maintaining perceptual consistency. The loss is:
\begin{equation}
\begin{aligned}
\mathcal{L}_{\text{recon}} &= (1-\lambda_s)\mathcal{L}_1(S_j, S^{\text{out}}_j) 
+ \lambda_s\mathcal{L}_{\text{D-SSIM}}(S_j, S^{\text{out}}_j), \\
\mathcal{L}_{\text{outpainted}} &= \mathcal{L}_{\text{recon}}(S_j, S^{\text{out}}_j) 
+ \lambda_{\text{perc}}\mathcal{L}_{\text{LPIPS}}(S_j, S^{\text{out}}_j),
\end{aligned}
\label{eq:loss_outpaint}
\end{equation}
where $S_j$ denotes the rendered wide-FOV image and $S^{\text{out}}_j$ is the generated outpainted image.

\section{Experiments}

\subsection{Experimental Setups}

\paragraph{Datasets and Evaluation Protocol.}
We evaluate on \textbf{Replica}~\cite{straub2019replica}, \textbf{ScanNet++}~\cite{yeshwanth2023scannet++}, and \textbf{Mip-NeRF 360}~\cite{barron2023mipnerf360}. 
Following~\cite{zhong2025taming,zhong2025empowering}, we adopt sparse-view settings with 3, 6, and 9 views for indoor datasets (\textbf{Replica} and \textbf{ScanNet++}), where the 9-view results are reported in the supplementary material, and 6 and 9 views for the unbounded \textbf{Mip-NeRF 360} dataset. 
For Replica and ScanNet++, we follow the view splits used in~\cite{zhong2025taming,zhong2025empowering}, while for Mip-NeRF 360 we use the original 6- and 9-view configurations. 
For ScanNet++, we additionally construct the 3- and 9-view setups to ensure suitable coverage for training and evaluation. 
Performance is evaluated using PSNR, SSIM~\cite{wang2004image}, and LPIPS~\cite{zhang2018unreasonable}.

\paragraph{Implementation Details.}
We adopt \textbf{MVGenMaster}~\cite{cao2025mvgenmaster} as the diffusion backbone to incorporate geometry-aware priors in a zero-shot manner. 
For fair comparison, image resolutions are standardized: \textbf{Replica} is set to $512 \times 384$, while \textbf{ScanNet++} and \textbf{Mip-NeRF 360} are set to $576 \times 384$. 
GaMO reconstructs a scene in approximately 8 minutes on a single RTX 4090 GPU, including diffusion outpainting and the final 3DGS optimization. 
Additional implementation details, hyperparameters, optimization settings, runtime breakdown, and extended results are provided in the supplementary material.

\begin{figure*}[t]
\centering
\begin{minipage}{\textwidth}
\centering
\small
\setlength{\tabcolsep}{4pt}
\renewcommand{\arraystretch}{1.1}

\captionof{table}{
\textbf{Quantitative comparison on Replica~\cite{straub2019replica} and ScanNet++~\cite{yeshwanth2023scannet++} with 3 and 6 input views.}
}
\label{tab:comparison_3v6v}

\resizebox{\textwidth}{!}{
\begin{tabular}{lccccccccccccc}
\toprule
& \multicolumn{3}{c}{\textsc{Replica (3 views)}} 
& \multicolumn{3}{c}{\textsc{Replica (6 views)}} 
& \multicolumn{3}{c}{\textsc{ScanNet++ (3 views)}} 
& \multicolumn{3}{c}{\textsc{ScanNet++ (6 views)}} & \\
\cmidrule(lr){2-4} \cmidrule(lr){5-7} \cmidrule(lr){8-10} \cmidrule(lr){11-13}

Method
& PSNR $\uparrow$ & SSIM $\uparrow$ & LPIPS $\downarrow$
& PSNR $\uparrow$ & SSIM $\uparrow$ & LPIPS $\downarrow$
& PSNR $\uparrow$ & SSIM $\uparrow$ & LPIPS $\downarrow$
& PSNR $\uparrow$ & SSIM $\uparrow$ & LPIPS $\downarrow$
& Run time \\
\midrule

3DGS & 20.39 & 0.818 & \cellcolor{yellow!25}0.154
& 24.74 & \cellcolor{yellow!25}0.862 & \cellcolor{orange!25}0.124
& 16.60 & 0.710 & 0.313
& 21.71 & \cellcolor{orange!25}0.818 & \cellcolor{orange!25}0.186 
& 2 min \\

FSGS & 20.84 & 0.815 & 0.172
& 23.91 & 0.846 & 0.145
& 16.62 & 0.690 & 0.359
& 21.69 & 0.801 & 0.298 
& 12 min \\

InstantSplat & \cellcolor{yellow!25}21.42 & 0.830 & \cellcolor{orange!25}0.137
& 23.09 & 0.849 & \cellcolor{yellow!25}0.141
& 16.72 & 0.720 & \cellcolor{yellow!25}0.312
& 21.19 & 0.811 & \cellcolor{yellow!25}0.193 
& 1 min \\

Difix3D+ & 19.47 & 0.783 & 0.194
& 21.86 & 0.811 & 0.188
& 15.64 & 0.659 & 0.346
& 20.62 & 0.764 & 0.244 
& 31 min \\

GenFusion & 22.34 & \cellcolor{yellow!25}0.833 & 0.172
& 23.98 & 0.855 & 0.142
& \cellcolor{yellow!25}17.97 & \cellcolor{yellow!25}0.725 & 0.354
& \cellcolor{yellow!25}21.96 & 0.808 & 0.218 
& 22 min \\

GuidedVD-3DGS$^\dagger$
& \textcolor{gray}{23.98} & \textcolor{gray}{0.848} & \textcolor{gray}{0.136}
& \textcolor{gray}{26.35} & \textcolor{gray}{0.872} & \textcolor{gray}{0.122}
& \textcolor{gray}{-} & \textcolor{gray}{-} & \textcolor{gray}{-}
& \textcolor{gray}{23.89} & \textcolor{gray}{0.850} & \textcolor{gray}{0.182} 
& - \\

GuidedVD-3DGS$^\ddagger$
& \cellcolor{red!25}25.26 & \cellcolor{orange!25}0.864 & 0.138
& \cellcolor{red!25}26.68 & \cellcolor{orange!25}0.880 & 0.133
& \cellcolor{orange!25}18.82 & \cellcolor{orange!25}0.720 & \cellcolor{orange!25}0.328
& \cellcolor{orange!25}22.98 & \cellcolor{yellow!25}0.815 & 0.204 
& 3h 20 min \\

Ours
& \cellcolor{orange!25}24.40 & \cellcolor{red!25}0.865 & \cellcolor{red!25}0.117
& \cellcolor{orange!25}26.40 & \cellcolor{red!25}0.882 & \cellcolor{red!25}0.104
& \cellcolor{red!25}20.06 & \cellcolor{red!25}0.759 & \cellcolor{red!25}0.265
& \cellcolor{red!25}23.41 & \cellcolor{red!25}0.835 & \cellcolor{red!25}0.181 
& 8 min \\

\bottomrule
\end{tabular}
}

\vspace{-3mm}
\begin{flushleft}
\scriptsize $^\dagger$ Reported in paper, $^\ddagger$ Our reproduction.
\end{flushleft}
\end{minipage}

\vspace{2mm}

\begin{minipage}{\textwidth}
\centering
\includegraphics[width=\linewidth]{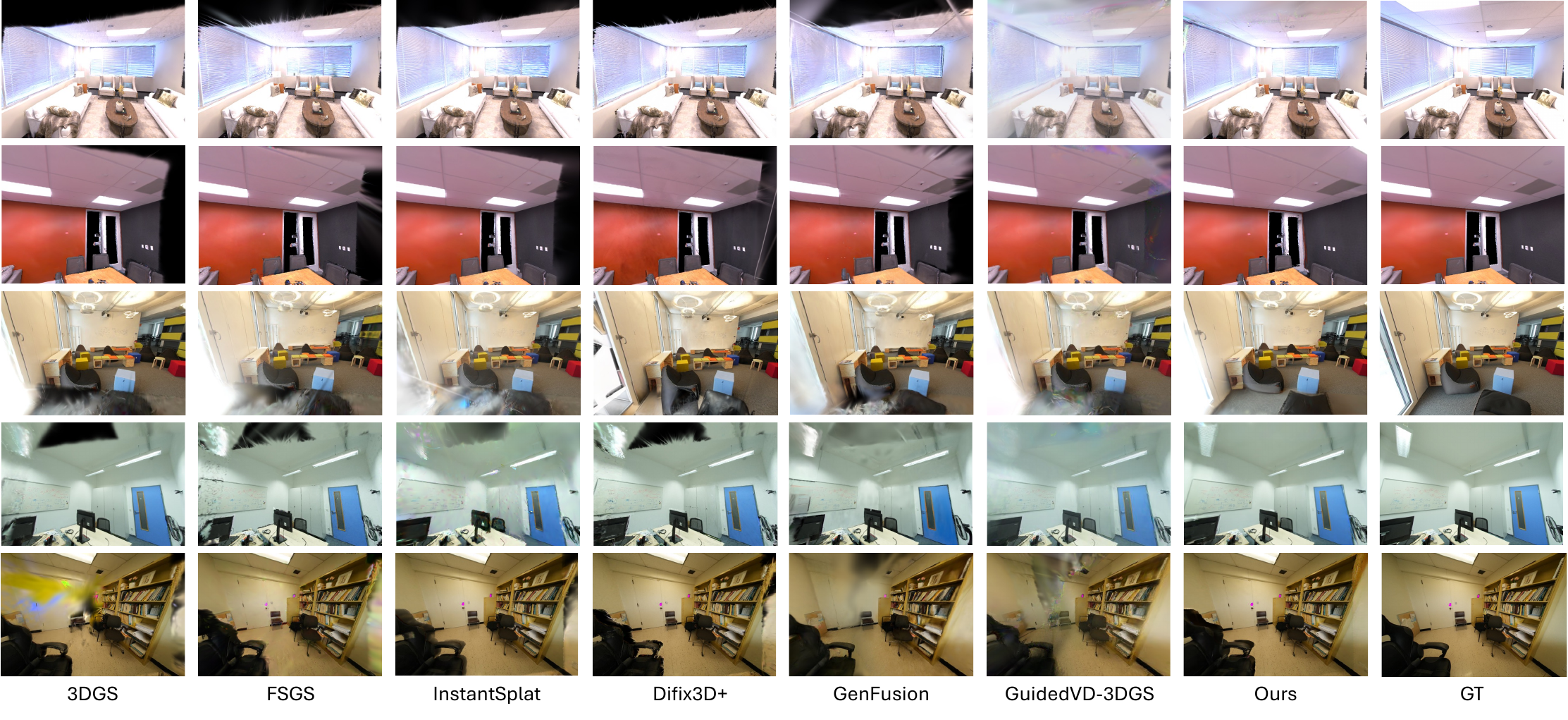}

\caption{
\textbf{Qualitative comparison on Replica and ScanNet++ under sparse 3-view and 6-view settings.}
Our method produces better coverage, improved geometric consistency, and fewer artifacts.
White boxes highlight challenging regions.
}
\label{fig:comparison}
\end{minipage}

\end{figure*}

\paragraph{Baselines.}
We compare against several state-of-the-art sparse-view reconstruction methods: 
(1) vanilla 3DGS; 
(2) \textbf{FSGS}~\cite{zhu2024fsgs}, using depth-guided Gaussian unpooling; 
(3) \textbf{InstantSplat}~\cite{fan2024instantsplat}, employing MASt3R priors and self-supervised bundle adjustment; 
(4) \textbf{Difix3D}~\cite{wu2025difix3d+}, applying single-step diffusion refinement; 
(5) \textbf{GenFusion}~\cite{wu2025genfusion}, integrating reconstruction and video diffusion via cyclical fusion; and 
(6) \textbf{GuidedVD-3DGS}~\cite{zhong2025taming}, leveraging video diffusion and evaluated using the authors' official implementation and settings. 
For fair comparison, all methods except InstantSplat use DUSt3R~\cite{wang2024dust3r} for initialization.


\subsection{Comparisons}
\begin{figure}[t]
\centering
\vspace{-2mm}

\begin{minipage}{\linewidth}
\centering
\scriptsize
\setlength{\tabcolsep}{4.5pt}
\renewcommand{\arraystretch}{0.9}

\captionof{table}{
\textbf{Quantitative comparison on MipNeRF360~\cite{barron2023mipnerf360} with 6 and 9 input views.}
}
\label{tab:mipnerf360_6_9_colored}

\begin{tabular}{lccc ccc}
\toprule
\multirow{2}{*}{Method} 
& \multicolumn{3}{c}{\textsc{Mip-NeRF 360 (6 views)}}
& \multicolumn{3}{c}{\textsc{Mip-NeRF 360 (9 views)}} \\
\cmidrule(lr){2-4}\cmidrule(lr){5-7}
 & PSNR $\uparrow$ & SSIM $\uparrow$ & LPIPS $\downarrow$
 & PSNR $\uparrow$ & SSIM $\uparrow$ & LPIPS $\downarrow$ \\
\midrule


3DGS 
& \cellcolor{yellow!25}{15.30} & \cellcolor{yellow!25}{0.342} & 0.459
& \cellcolor{orange!25}16.29 & \cellcolor{yellow!25}0.383 & \cellcolor{yellow!25}0.385 \\

FSGS 
& 14.76 & 0.326 & 0.532
& 15.36 & 0.374 & 0.502 \\

InstantSplat 
& \cellcolor{orange!25}{15.91} & \cellcolor{orange!25}{0.388} & \cellcolor{yellow!25}0.443
& \cellcolor{yellow!25}16.10 & \cellcolor{orange!25}{0.433} & 0.419 \\

Difix3D+ 
& 14.94 & 0.308 & \cellcolor{red!25}{0.419}
& 16.04 & 0.368 & \cellcolor{red!25}{0.371} \\

GenFusion & 16.40 & 0.384 & 0.487 & 17.55 & 0.435 & 0.409 \\

GuidedVD-3DGS
& 13.89 & 0.273 & 0.640
& 15.77 & 0.386 & 0.418 \\

Ours 
& \cellcolor{red!25}{16.74} & \cellcolor{red!25}{0.393} & \cellcolor{orange!25}{0.436}
& \cellcolor{red!25}{17.56} & \cellcolor{red!25}{0.448} & \cellcolor{orange!25}{0.381} \\

\bottomrule
\end{tabular}

\end{minipage}

\vspace{3mm}

\begin{minipage}{\linewidth}
\centering
\includegraphics[width=\linewidth]{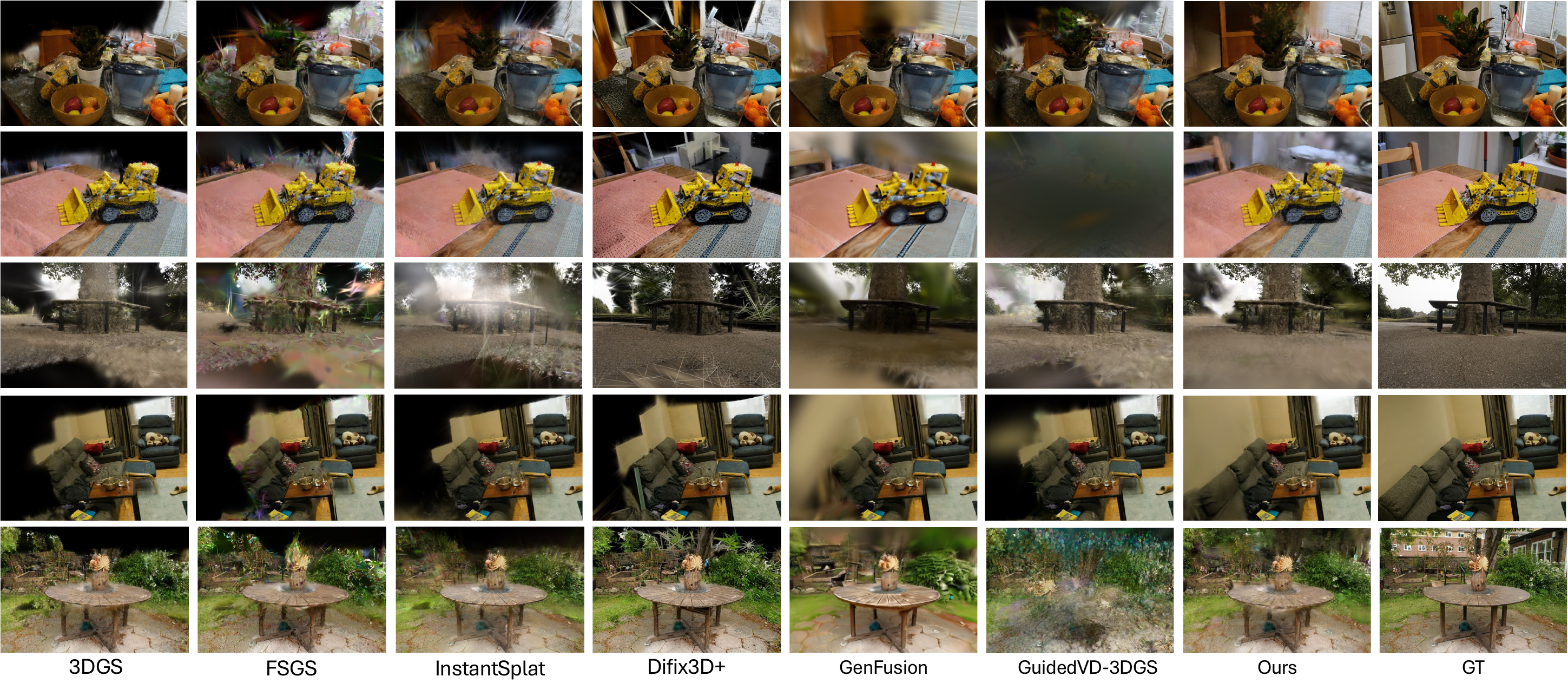}

\captionof{figure}{
\textbf{Qualitative comparison on MipNeRF360~\cite{barron2023mipnerf360}under sparse 6-view and 9-view settings.}
Large-scale outdoor scenes with wide baselines make sparse-view reconstruction challenging.
Our method achieves better coverage, improved geometric consistency, and fewer artifacts than prior methods.
White boxes highlight challenging regions.
}
\label{fig:mipnerf360}
\end{minipage}

\end{figure}

\noindent\textbf{Quantitative Results.} 
Tab.~\ref{tab:comparison_3v6v} and Tab.~\ref{tab:mipnerf360_6_9_colored} report comparisons on Replica, ScanNet++, and Mip-NeRF 360 under sparse-view settings. Our method consistently achieves strong performance across all datasets. On Replica and ScanNet++, we obtain the best SSIM and lowest LPIPS in most settings, indicating superior structural consistency and perceptual quality. On the challenging Mip-NeRF 360 benchmark, our method achieves the best PSNR and SSIM for both 6-view and 9-view inputs while maintaining competitive LPIPS. In addition, GaMO is highly efficient, achieving up to $25\times$ speedup over the diffusion-based GuidedVD-3DGS~\cite{zhong2025taming}, reducing the reconstruction time to under 10 minutes.

\noindent\textbf{Qualitative Results.}
Fig.~\ref{fig:comparison} and Fig.~\ref{fig:mipnerf360} show visual comparisons on representative scenes. Our method produces more complete reconstructions by effectively addressing the key challenges in sparse-view reconstruction: reducing black holes in unobserved regions, minimizing rendering artifacts, and improving geometric consistency. These results demonstrate the effectiveness of GaMO in generating high-quality outpainted views that enhance 3D reconstruction.

\subsection{Ablation Studies}

\begin{figure}[t]
    \centering
    \footnotesize
    \begin{minipage}[t]{0.54\textwidth}
        \vspace{0pt} 
        \captionof{table}{\textbf{Quantitative ablation of blending components.} We evaluate the impact of augmented conditioning (Aug.), hard (H.)/soft (S.) mask blending, and noise resampling (N.). P, S, L denote PSNR$\uparrow$, SSIM$\uparrow$, and LPIPS$\downarrow$, respectively. Row 5 represents our full model.}
        \label{tab:ablation_1} 
        \resizebox{\linewidth}{!}{
            \setlength{\tabcolsep}{2pt}
            \begin{tabular}{l cccc ccc ccc}
                \toprule
                & \multicolumn{4}{c}{Components} & \multicolumn{3}{c}{Outpainted} & \multicolumn{3}{c}{Novel View} \\
                \cmidrule(lr){2-5}\cmidrule(lr){6-8}\cmidrule(lr){9-11}
                \# & {Aug.} & {H.} & {S.} & {N.} & {P} & {S} & {L} & {P} & {S} & {L} \\
                \midrule
                1 & & & & & 18.97 & .776 & .210 & 22.37 & .821 & .197 \\
                2 & \checkmark & & & & 19.11 & .779 & .207 & 22.53 & .822 & .192 \\
                3 & \checkmark & \checkmark & & & 19.77 & .797 & .199 & 23.52 & .839 & .174 \\
                4 & \checkmark & & \checkmark & \checkmark & 19.37 & .797 & .196 & 23.22 & \textbf{.840} & .173 \\
                \rowcolor{black!10}
                5* & \checkmark & \checkmark & \checkmark & \checkmark & \textbf{20.01} & \textbf{.800} & \textbf{.190} & \textbf{23.53} & .839 & \textbf{.172} \\
                \bottomrule
            \end{tabular}
        }
    \end{minipage}
    \hfill
    \begin{minipage}[t]{0.44\textwidth}
        \vspace{2.2mm} 
        \centering
        \includegraphics[width=1\linewidth]{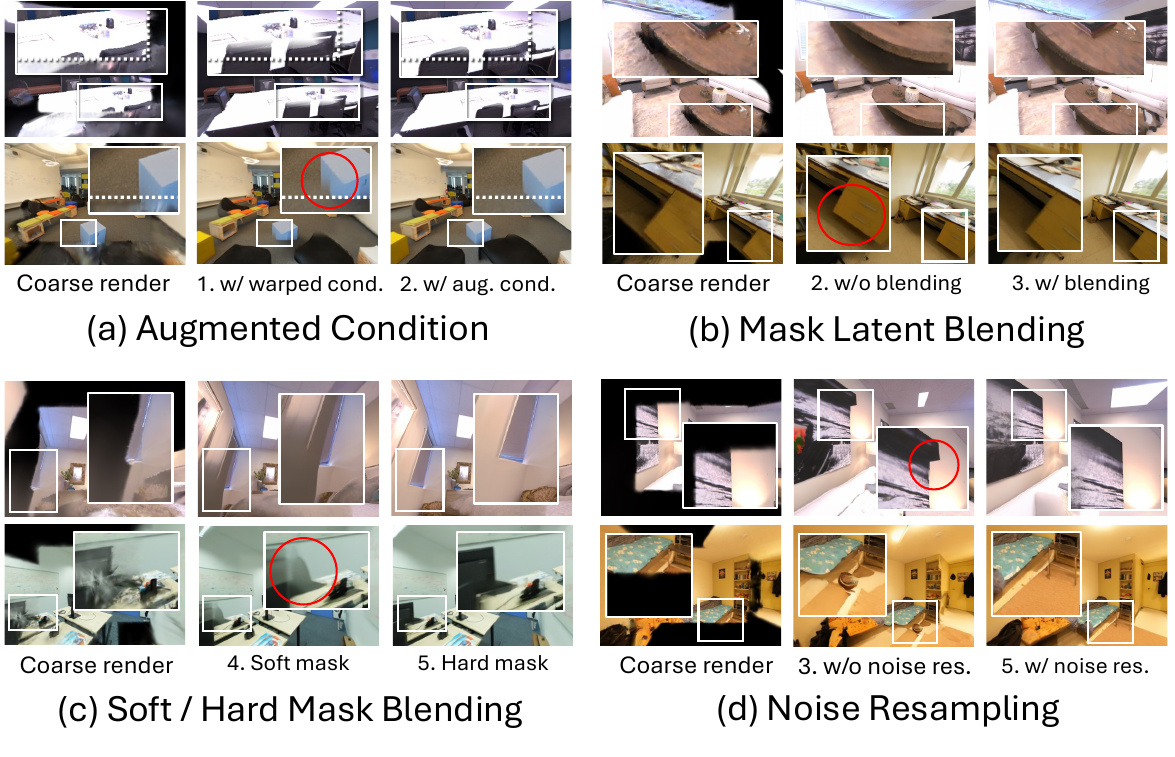}
        \vspace{-5mm} 
        \caption{\textbf{Qualitative ablation.} (a) Aug. cond. aligns content. (b-c) Mask blending improves geometry. (d) Noise resampling removes seams.}
        \label{fig:ablation_1}
    \end{minipage}
\end{figure}

\label{sec:ablation}
We conduct comprehensive ablation studies on Replica and ScanNet++ datasets with 6 input views. To separately assess the outpainted view quality and novel view synthesis performance, we center crop the input images to 0.6× of their original size. For each ablation, we provide quantitative results and visual comparisons, where numbered configurations (e.g., ``1.'') in figures correspond to table rows, and letters (e.g., (a), (b)) denote different visual comparison aspects.

\paragraph{Latent Blending Strategies.}
Tab.~\ref{tab:ablation_1} and Fig.~\ref{fig:ablation_1} present ablation results on our latent blending design. Augmenting the warped features with downscaled reference RGB and CCM (rows 1-2) prevents incorrect hallucinations in known regions (a). Mask latent blending (rows 2-3) prevents severe geometric misalignment (b, red circle) and improves PSNR by 0.66 dB. Hard masking (rows 4-5) produces sharper boundaries (c) with 0.64 dB gain over soft masking. Finally, noise resampling (rows 3 vs. 5) reduces blending artifacts by 0.24 dB, generating more coherent results (d).

\begin{figure}[t]
    \centering
    \footnotesize
    \begin{minipage}[t]{0.56\textwidth}
        \vspace{0pt} 
        \captionof{table}{\textbf{Ablation on mask blending scheduling.} Comparison of blending at different timesteps: $t_k$ (single-step), $t_{1\to t_N}$ (multi-step), All (every step), and IMS (Iterative Mask Scheduling). Row 4* is our full method. \textit{Time (s)} is total generation time.}
        \label{tab:ablation_2} 
        \resizebox{\linewidth}{!}{
            \setlength{\tabcolsep}{2pt}
            \begin{tabular}{l cccc cccc ccc}
                \toprule
                & \multicolumn{4}{c}{Scheduling} & \multicolumn{4}{c}{Outpainted} & \multicolumn{3}{c}{Novel View} \\
                \cmidrule(lr){2-5}\cmidrule(lr){6-9}\cmidrule(lr){10-12}
                \# & {$t_k$} & {$t_{1\!\to\!t_N}$} & {All} & {IMS} & {P$\uparrow$} & {S$\uparrow$} & {L$\downarrow$} & {T(s)$\downarrow$} & {P$\uparrow$} & {S$\uparrow$} & {L$\downarrow$} \\
                \midrule
                1 & \checkmark & & & & 20.09 & .804 & .198 & \textbf{85} & 23.38 & .837 & .176 \\
                2 & & \checkmark & & & 19.85 & .799 & .173 & 93 & 23.53 & .839 & .179 \\
                3 & & & \checkmark & & \textbf{20.31} & \textbf{.809} & .201 & 167 & \textbf{23.67} & \textbf{.842} & .178 \\
                \rowcolor{black!10}
                4* & & \checkmark & & \checkmark & 20.07 & .801 & \textbf{.169} & 93 & 23.65 & .839 & \textbf{.171} \\
                \bottomrule
            \end{tabular}
        }
    \end{minipage}
    \hfill
    \begin{minipage}[t]{0.42\textwidth}
        \vspace{2.2mm} 
        \centering
        \includegraphics[width=1\linewidth]{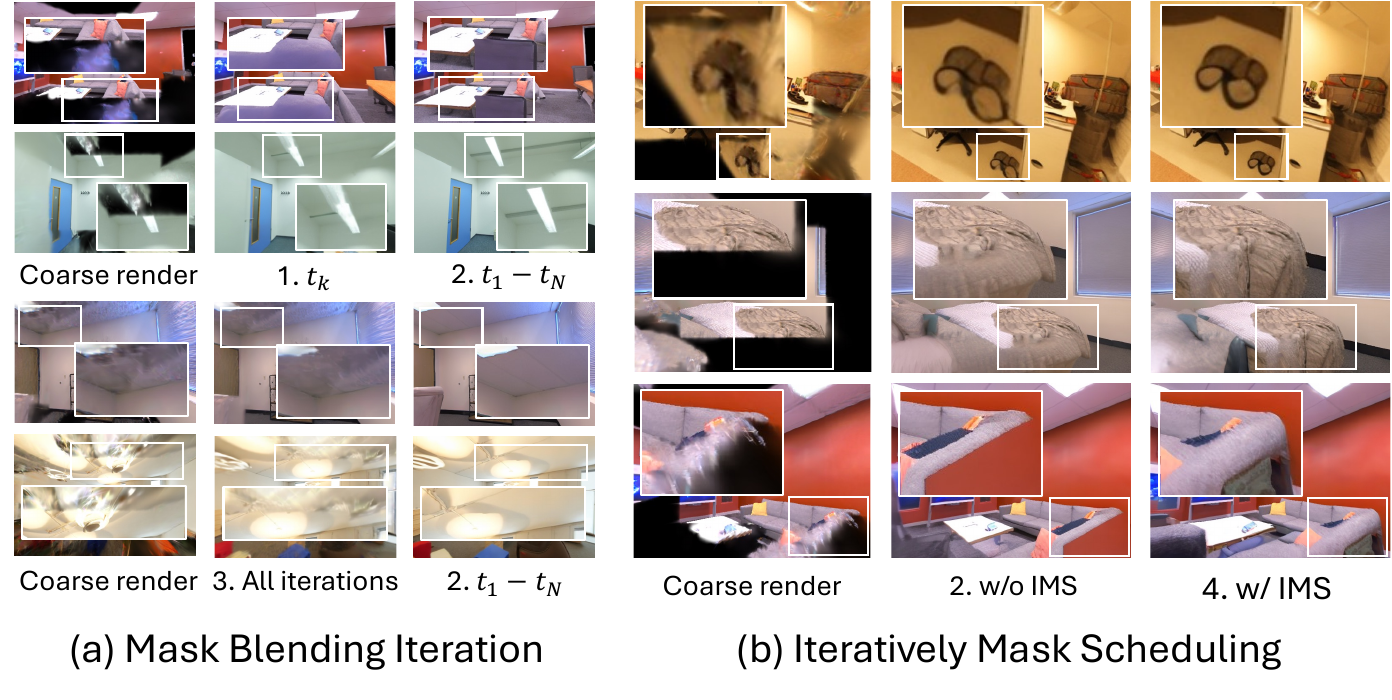}
        \vspace{-5mm} 
        \caption{\textbf{Qualitative ablation.} (a) Blending timesteps impact. (b) IMS progressively improves detail coherence and alignment.}
        \label{fig:ablation_2}
    \end{minipage}
\end{figure}

\paragraph{Mask Blending Scheduling.}
Tab.~\ref{tab:ablation_2} and Fig.~\ref{fig:ablation_2} present ablation results on mask blending scheduling strategies. Single-step blending (row 1) is insufficient as coarse geometry is easily washed out during denoising, while multi-step blending (row 2) better preserves geometric cues (a-top). Blending at every step (row 3) achieves slightly higher PSNR/SSIM but causes blurred boundaries (a-bottom) and increases denoising time, making the range-based approach preferable. Finally, Iterative Mask Scheduling (rows 2 vs. 4) substantially improves perceptual quality through progressive mask dilation, providing better geometric guidance and smoother transitions for more coherent details (b).

\begin{figure}[t]
    \centering
    \footnotesize
    \begin{minipage}[t]{0.44\textwidth}
        \vspace{0pt} 
        \captionof{table}{\textbf{Ablation on 3DGS refinement.} Impact of point re-init. and perceptual loss on reconstruction quality. Row 3* is our full method.}
        \label{tab:ablation_3} 
        \resizebox{\linewidth}{!}{
            \setlength{\tabcolsep}{3pt}
            \begin{tabular}{l cc ccc}
                \toprule
                & \multicolumn{2}{c}{Components} & \multicolumn{3}{c}{Novel View} \\
                \cmidrule(lr){2-3}\cmidrule(lr){4-6}
                \# & {Re-init.} & {Percep.} & {P$\uparrow$} & {S$\uparrow$} & {L$\downarrow$} \\
                \midrule
                1 & & \checkmark & 24.80 & .860 & .140 \\
                2 & \checkmark & & \textbf{25.14} & .857 & .139 \\
                \rowcolor{black!10}
                3* & \checkmark & \checkmark & 24.93 & \textbf{.861} & \textbf{.135} \\
                \bottomrule
            \end{tabular}
        }
    \end{minipage}
    \hfill
    \begin{minipage}[t]{0.54\textwidth}
        \vspace{2.2mm} 
        \centering
        \includegraphics[width=1\linewidth]{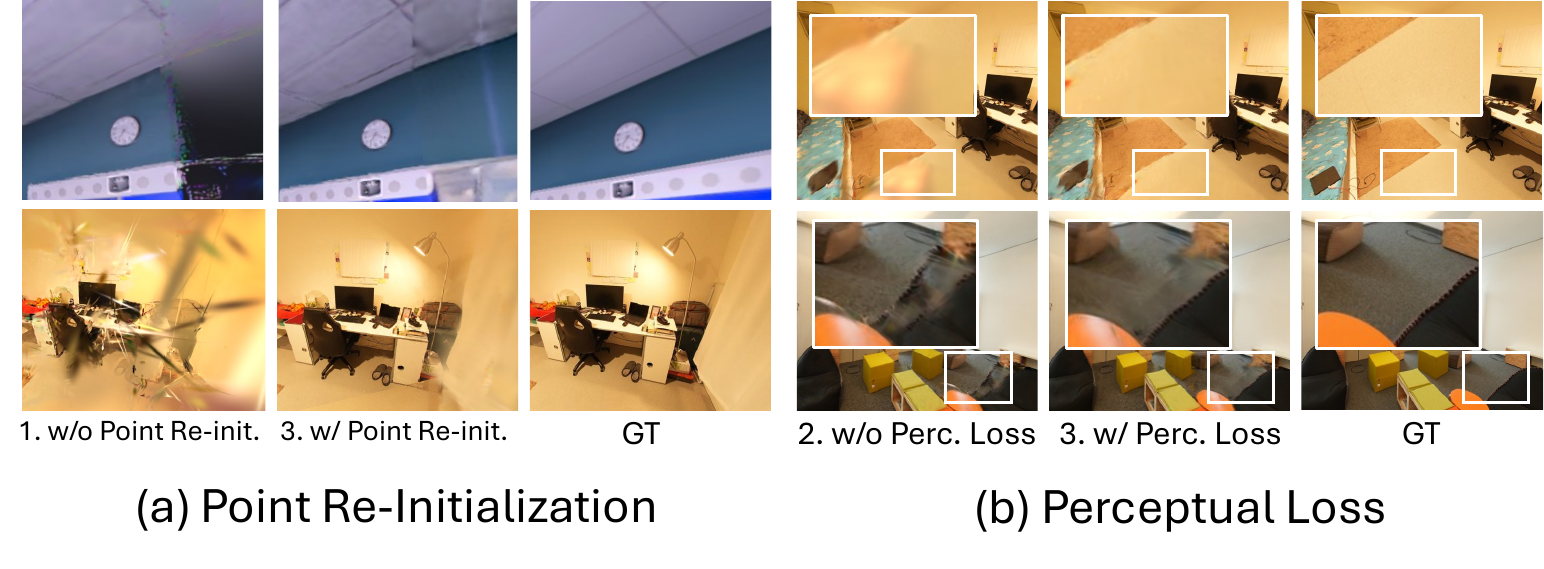}
        \vspace{-5mm} 
        \caption{\textbf{Qualitative ablation.} (a) Re-init. enables Gaussian generation in new regions. (b) Perceptual loss recovers sharp details.}
        \label{fig:ablation_3}
    \end{minipage}
\end{figure}

\paragraph{3DGS Refinement Components.}
Tab.~\ref{tab:ablation_3} and Fig.~\ref{fig:ablation_3} present ablation results on 3DGS refinement components. Point cloud re-initialization using outpainted views (rows 1 vs. 3) enables the successful generation of Gaussian points in outpainted regions (a). Perceptual loss (rows 2 vs. 3) effectively fills holes and reduces artifacts by providing better gradient guidance for outpainted regions (b), producing cleaner and more realistic renderings.

\section{Conclusion}
\label{sec:conclusion}
We present \textit{GaMO}, which formulates novel view generation as an outpainting problem for sparse-view 3D reconstruction. By extending existing views instead of generating new perspectives, our approach preserves geometric consistency while expanding spatial coverage, effectively reducing holes and artifacts. Experiments show consistent improvements over prior methods with superior reconstruction quality and several tens of times faster runtime. GaMO also demonstrates strong zero-shot generalization, highlighting outpainting as an efficient and principled paradigm for sparse-view reconstruction.

\paragraph{Limitations.}
GaMO cannot recover content fully occluded from all input views. Its performance also depends on input view distribution, where clustered or misaligned views may degrade results. Future work could explore adaptive outpaint scaling and hybrid strategies for more challenging cases.

\section*{Acknowledgements}
This research was funded by the National Science and Technology Council, Taiwan, under Grants NSTC 112-2222-E-A49-004-MY2 and 113-2628-E-A49-023-. The authors are grateful to Google, NVIDIA, and MediaTek Inc. for their generous donations. Yu-Lun Liu acknowledges the Yushan Young Fellow Program by the MOE in Taiwan.

%
%
\bibliographystyle{splncs04}
\bibliography{main}

\clearpage

\appendix
\section{Overview}

This supplementary material provides additional details and analyses complementing the main paper. It is organized as follows:

\begin{enumerate}

\item \textbf{Generation-based Comparison (Sec.~\ref{sec:gen_comp})}. 
We compare our method with a diffusion-based novel view generation approach, \textbf{SEVA}~\cite{zhou2025stable}, to highlight the differences between generation-based pipelines and our outpainting-based sparse-view reconstruction framework.

\item \textbf{Implementation Details (Sec.~\ref{sec:impl})}. 
We provide detailed implementation information for GaMO, including preprocessing, diffusion inference settings, mask generation, and integration with 3D Gaussian Splatting.

\item \textbf{Outpainting Comparison with Multi-View Diffusion Models (Sec.~\ref{sec:outpaint_comp})}. 
We evaluate different diffusion backbones for the outpainting stage and analyze their impact on reconstruction quality and geometric consistency.

\item \textbf{Iterative Mask Scheduling (Sec.~\ref{sec:ims})}. 
We present additional details and visualizations of the proposed iterative mask scheduling strategy and its effect on boundary coherence and generation stability.

\item \textbf{Additional Quantitative Comparisons (Sec.~\ref{sec:quant})}. 
We report extended quantitative evaluations across datasets and sparse-view settings (3, 6, and 9 views).

\item \textbf{Additional Qualitative Comparisons (Sec.~\ref{sec:qual})}. 
We provide additional visual comparisons illustrating reconstruction quality and scene completeness.

\item \textbf{Runtime Analysis (Sec.~\ref{sec:runtime})}. 
We analyze the computational efficiency of GaMO and compare runtime with existing diffusion-based reconstruction pipelines.

\item \textbf{Failure Cases (Sec.~\ref{sec:failure})}. 
We present representative failure cases and discuss limitations of the proposed approach.

\item \textbf{Baseline Implementation Details (Sec.~\ref{sec:baseline})}. 
We describe the implementation details of all baseline methods, including training configurations, initialization strategies, and evaluation protocols.

\item \textbf{Per-Scene Quantitative Results (Sec.~\ref{sec:per_scene})}. 
We report detailed PSNR, SSIM, and LPIPS results for every scene across Replica~\cite{straub2019replica}, ScanNet++~\cite{yeshwanth2023scannet++}, and Mip-NeRF 360~\cite{barron2023mipnerf360}.

\end{enumerate}

In addition, we provide an interactive HTML visualization (\texttt{main.html}) showing rendered videos along novel-view trajectories across scenes, enabling qualitative inspection of reconstruction quality beyond the input viewpoints.

\section{Generation-based Comparison}
\label{sec:gen_comp}

\begin{figure}[ht]
    \centering
    \footnotesize
    \begin{minipage}[t]{0.48\textwidth}
        \vspace{0pt}
        \captionof{table}{\textbf{Comparison with the diffusion-based novel view generation method SEVA.} 
        We report PSNR$\uparrow$, SSIM$\uparrow$, and LPIPS$\downarrow$.}
        \label{tab:seva_comparison}
        \resizebox{\linewidth}{!}{
        \setlength{\tabcolsep}{4pt}
        \renewcommand{\arraystretch}{1.15}
        \begin{tabular}{l!{\vrule}ccc!{\vrule}ccc}
            \toprule
            & \multicolumn{3}{c!{\vrule}}{\textbf{Replica (3v)}}
            & \multicolumn{3}{c}{\textbf{ScanNet++ (3v)}} \\
            \cmidrule(lr){2-4}\cmidrule(lr){5-7}
            & PSNR & SSIM & LPIPS & PSNR & SSIM & LPIPS \\
            \midrule
            SEVA & 18.52 & 0.677 & 0.193 & 15.79 & 0.631 & 0.343 \\
            Ours & \textbf{25.40} & \textbf{0.864} & \textbf{0.117}
                 & \textbf{20.01} & \textbf{0.765} & \textbf{0.266} \\
            \midrule
            & \multicolumn{3}{c!{\vrule}}{\textbf{Replica (6v)}}
            & \multicolumn{3}{c}{\textbf{ScanNet++ (6v)}} \\
            \cmidrule(lr){2-4}\cmidrule(lr){5-7}
            & PSNR & SSIM & LPIPS & PSNR & SSIM & LPIPS \\
            \midrule
            SEVA & 18.64 & 0.691 & 0.206 & 16.93 & 0.640 & 0.276 \\
            Ours & \textbf{25.84} & \textbf{0.877} & \textbf{0.109}
                 & \textbf{23.41} & \textbf{0.835} & \textbf{0.181} \\
            \bottomrule
        \end{tabular}
        }
    \end{minipage}
    \hfill
    \begin{minipage}[t]{0.50\textwidth}
        \vspace{2mm}
        \centering
        \includegraphics[width=\linewidth]{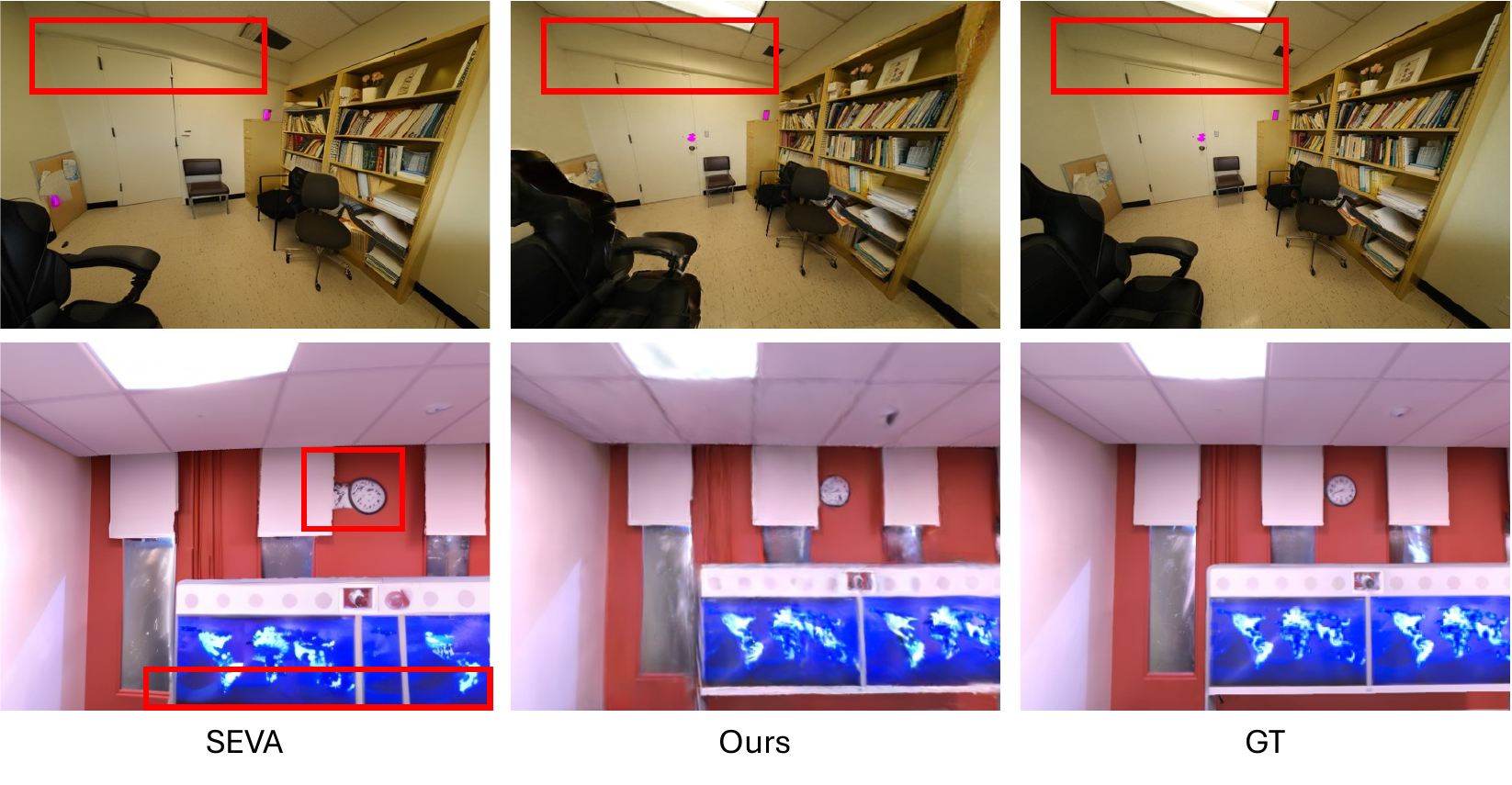}
        \vspace{-6mm}
        \caption{\textbf{Qualitative comparison with SEVA.} 
While SEVA produces visually plausible results, it often lacks pixel-level accuracy and geometric consistency.}
        \label{fig:ablation_1}
    \end{minipage}
\end{figure}

To further compare with approaches designed for direct novel view generation, we evaluate our method against the diffusion-based method SEVA~\cite{zhou2025stable}. As shown in Table~\ref{tab:seva_comparison}, our approach consistently outperformed SEVA across both datasets and view settings. In particular, our method achieved significantly higher PSNR and SSIM while reducing LPIPS, indicating more accurate reconstruction and better perceptual quality.

Qualitative results in Fig.~\ref{fig:ablation_1} further illustrated these differences. While SEVA produced visually plausible results, it often exhibited inaccurate pixel-level details and geometrically inconsistent structures. In contrast, our method preserved accurate details and more coherent scene geometry.

\vspace{-10pt}

\vspace{-10pt}

\section{Implementation Details}
\label{sec:impl}

For coarse initialization, we train 3DGS for 10,000 iterations with $\lambda_s = 0.2$ and opacity threshold $\eta_{\text{mask}} = 0.6$. 

For outpainting, we use the multi-view diffusion model~\cite{cao2025mvgenmaster} with focal-length scaling $S_k \in [0.5, 0.7]$, adjusted per scene depending on scene scale. We use DDIM sampling~\cite{song2021denoising} with $T = 50$ steps, and perform latent blending at timesteps $t_1 = 0.7T$, $t_2 = 0.5T$, $t_3 = 0.3T$ with noise resampling $R = 3$. Input and outpainted views share the same resolution (differing only in FOV), with dimensions set as multiples of 64. Before refinement, we alpha-blend downscaled inputs at the center. 

For refinement, we optimize 3DGS for 3,000 iterations (3 views) or 7,000 iterations (6/9 views) with $\lambda_{\text{perc}} = 0.1$, alternating supervision between input and outpainted views. In addition, we make minor adjustments to several 3DGS refinement hyperparameters based on scene characteristics to ensure stable optimization. We will release all code, configurations, and scripts used in our experiments. All experiments are conducted on a single NVIDIA RTX 4090 GPU.

\section{Outpainting Comparison Using Multi-View Diffusion Models}
\label{sec:outpaint_comp}

\vspace{-10pt}
\begin{figure}[ht]
    \centering
    \includegraphics[width=\linewidth]{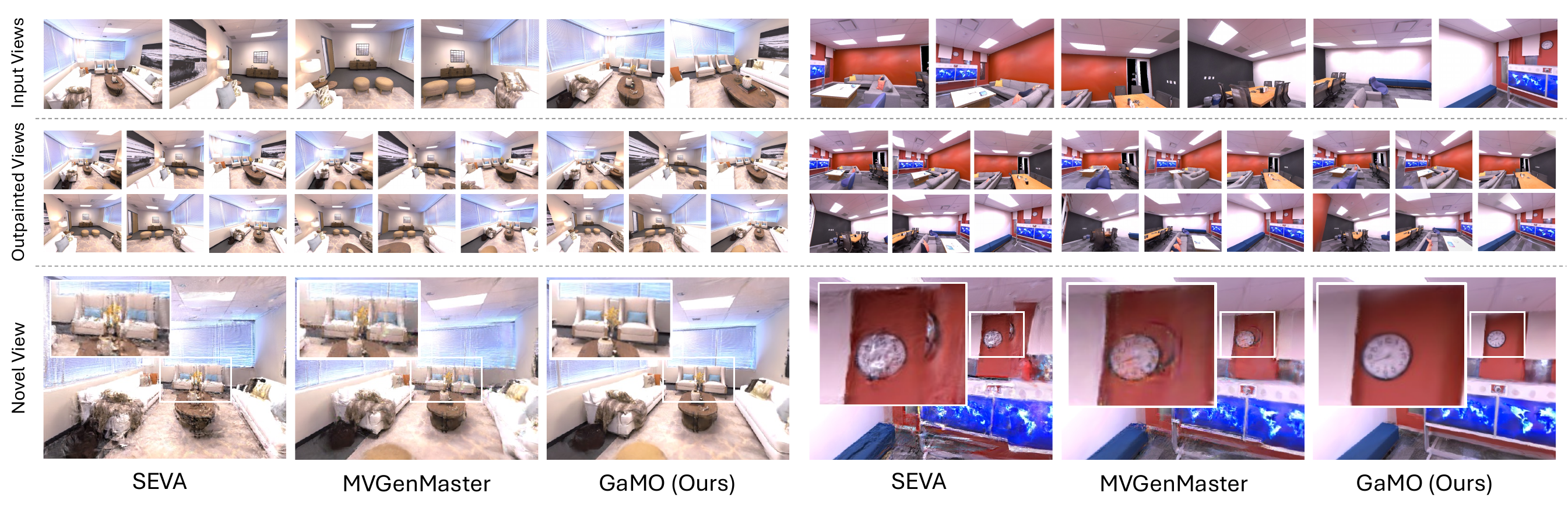}
    \vspace{-6mm}
    \caption{
        \textbf{Comparison of outpainting using adapted multi-view diffusion models.}
        Top: input views. 
        Middle: outpainted views generated by adapted SEVA~\cite{zhou2025stable}, MVGenMaster~\cite{cao2025mvgenmaster}, and our GaMO. 
        Bottom: novel views after 3DGS refinement using the generated outpainted views.
        Adapted multi-view diffusion models suffer from multi-view inconsistency, resulting in noisy reconstructions, while our method produces consistent outpainted views that improve reconstruction quality.
    }
    \label{fig:mv_diffusion_comparison}
\end{figure}

We compare our method against adapted multi-view diffusion models for outpainting. Specifically, we adapt SEVA~\cite{zhou2025stable} and MVGenMaster~\cite{cao2025mvgenmaster} 
by modifying the camera intrinsics to generate outpainted versions of the input views with 
extended FOV. These outpainted input views are then used to train 3DGS for improved novel 
view synthesis.

As shown in Fig.~\ref{fig:mv_diffusion_comparison}, SEVA produces highly noisy novel views after 3DGS refinement due to severe multi-view inconsistency caused by lack of geometric constraints. While MVGenMaster incorporates additional geometric mechanisms (e.g., multi-view conditioning), it still suffers from inconsistency issues that introduce artifacts in the refined reconstruction. In contrast, our GaMO effectively addresses the multi-view inconsistency problem, providing consistent outpainted views across multiple viewpoints that successfully refine 3DGS quality without introducing additional noise or artifacts.

\begin{wrapfigure}{r}{0.45\textwidth}
    \vspace{-12mm}
    \centering
    \includegraphics[width=\linewidth]{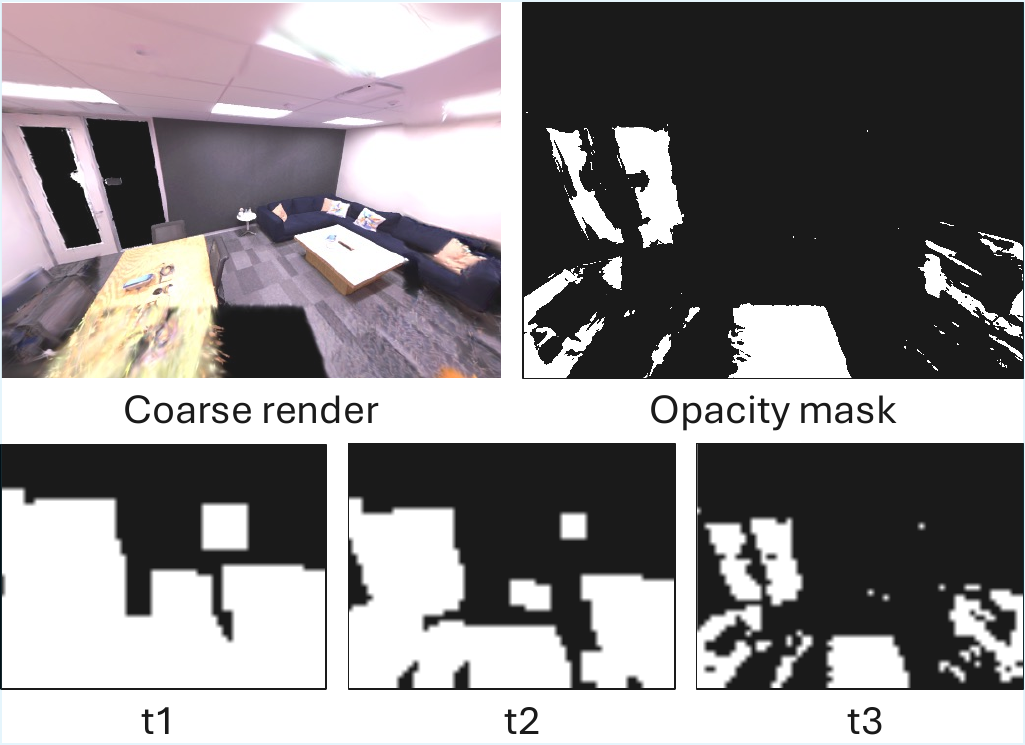}
    \vspace{-8mm}
    \caption{
    \textbf{Iterative Mask Scheduling visualization.}
    Top: coarse render and opacity mask derived from the coarse 3D initialization.
    Bottom: progressive mask shrinking during denoising at timesteps $t = 35, 25, 15$, with 2, 1, and 0 dilation iterations, respectively.
    }
    \label{fig:ims_vis}
    \vspace{-12mm}
\end{wrapfigure}
\section{Iterative Mask Scheduling Implementation}
\label{sec:ims}

To maximize the utilization of coarse geometry during outpainting while preserving the generative diversity of the diffusion model, we introduce an \textbf{Iterative Mask Scheduling (IMS)} strategy. IMS dynamically adjusts the mask region throughout the denoising process, allowing the diffusion model to first freely hallucinate missing regions and later progressively align the generated content with the coarse 3D initialization.

\noindent\textbf{Design Rationale.} 
As demonstrated in the ablation studies in the main paper (Tab. 3), we found that applying mask latent blending at specific denoising steps yields significantly better results than continuous blending throughout the entire denoising process. Based on these findings, we strategically select three representative timesteps corresponding to the early, middle, and late stages of denoising to participate in the latent blending process. At each stage, we employ progressively shrinking mask sizes to control the degree of interference with the denoising process: larger masks in early stages allow more freedom for generation, while smaller masks in later stages enforce stronger alignment with coarse geometry.

\noindent\textbf{Implementation Details.}
As illustrated in Fig.~\ref{fig:ims_vis}, we generate three mask levels through morphological dilation:
\begin{equation}
\mathcal{M}_{\text{latent}}^{(k)} = \text{Dilate}(\mathcal{M}_{\text{base}}^{\downarrow}, \text{kernel}=5, \text{iterations}=\frac{k-15}{10}),
\end{equation}
where $\mathcal{M}_{\text{base}}^{\downarrow}$ denotes the downsampled base mask from the coarse geometry opacity map, aligned to the latent space resolution of $64 \times 48$ via adaptive max pooling. The $\text{Dilate}(\cdot)$ operation applies iterative max pooling with $5 \times 5$ kernel to expand the masked region. During the denoising process from $t = 50$ to $t = 0$, we apply $\mathcal{M}_{\text{latent}}^{(35)}$ at $t = 35$, $\mathcal{M}_{\text{latent}}^{(25)}$ at $t = 25$, and $\mathcal{M}_{\text{latent}}^{(15)}$ at $t = 15$, as visualized in Fig.~\ref{fig:ims_vis}. This staged approach balances generative freedom with geometric consistency, as validated by our ablation experiments.

\section{More Quantitative Comparison}
\label{sec:quant}
We provide additional quantitative results on Replica~\cite{straub2019replica} and ScanNet++~\cite{yeshwanth2023scannet++} datasets with varying numbers of input views (3, 6, and 9 views), as shown in Tab.~\ref{tab:replica_3_6_9} and Tab.~\ref{tab:scannetpp_3_6_9}. We focus our comparison on 3DGS~\cite{kerbl20233d} and GuidedVD-3DGS~\cite{zhong2025taming}, a competitive state-of-the-art diffusion-based method.

\noindent\textbf{Evaluation Protocol.}
For Replica, we follow the evaluation protocol from~\cite{zhong2025taming} for all three view settings. For ScanNet++, the 6-view setting follows~\cite{zhong2025taming}, while the 3-view and 9-view settings use manually selected views to maximize spatial coverage. All methods use DUSt3R~\cite{wang2024dust3r} for point cloud initialization.

\noindent\textbf{Results.}
Our method consistently outperforms baselines across most metrics and view settings. On Replica, we achieve the best SSIM and LPIPS scores across all view counts. On ScanNet++, we obtain superior performance across all metrics in all view settings. Notably, our method maintains competitive quality with GuidedVD-3DGS~\cite{zhong2025taming} while being significantly faster (approximately 6-9 minutes vs. 3+ hours).

\begin{table*}[h!]
\centering
\scriptsize
\setlength{\tabcolsep}{3pt}
\caption{\textbf{Quantitative comparison on Replica~\cite{straub2019replica} with 3, 6, and 9 input views.}}
\label{tab:replica_3_6_9}
\vspace{-3mm}
\begin{tabular}{l!{\vrule}ccc!{\vrule}ccc!{\vrule}ccc}
\toprule
\multirow{2}{*}{Method} &
\multicolumn{3}{c!{\vrule}}{Replica (3 views)} &
\multicolumn{3}{c!{\vrule}}{Replica (6 views)} &
\multicolumn{3}{c}{Replica (9 views)} \\
\cmidrule(lr){2-4} \cmidrule(lr){5-7} \cmidrule(lr){8-10}
& PSNR & SSIM & LPIPS & PSNR & SSIM & LPIPS & PSNR & SSIM & LPIPS \\
\midrule
3DGS~\cite{kerbl20233d}
    & 20.39 & 0.818 & 0.154
    & 24.41 & 0.862 & \cellcolor{orange!25}0.124
    & 26.09 & 0.890 & \cellcolor{orange!25}0.100 \\
GuidedVD-3DGS~\cite{zhong2025taming}
    & \cellcolor{red!25}25.26 & \cellcolor{orange!25}0.864 & \cellcolor{orange!25}0.138
    & \cellcolor{red!25}26.68 & \cellcolor{orange!25}0.880 & 0.133
    & \cellcolor{red!25}28.08 & \cellcolor{orange!25}0.901 & 0.108 \\
Ours
    & \cellcolor{orange!25} 24.40 & \cellcolor{red!25}0.865 & \cellcolor{red!25}0.117
    & \cellcolor{orange!25}26.40 & \cellcolor{red!25}0.882 & \cellcolor{red!25}0.104
    & \cellcolor{orange!25} 27.58 & \cellcolor{red!25} 0.902 & \cellcolor{red!25} 0.096 \\
\bottomrule
\end{tabular}
\end{table*}

\begin{table*}[h!]
\centering
\scriptsize
\setlength{\tabcolsep}{3pt}
\caption{\textbf{Quantitative comparison on ScanNet++~\cite{yeshwanth2023scannet++} with 3, 6, and 9 input views.}}
\label{tab:scannetpp_3_6_9}
\vspace{-3mm}
\begin{tabular}{l!{\vrule}ccc!{\vrule}ccc!{\vrule}ccc}
\toprule
\multirow{2}{*}{Method} &
\multicolumn{3}{c!{\vrule}}{ScanNet++ (3 views)} &
\multicolumn{3}{c!{\vrule}}{ScanNet++ (6 views)} &
\multicolumn{3}{c}{ScanNet++ (9 views)} \\
\cmidrule(lr){2-4} \cmidrule(lr){5-7} \cmidrule(lr){8-10}
& PSNR & SSIM & LPIPS & PSNR & SSIM & LPIPS & PSNR & SSIM & LPIPS \\
\midrule
3DGS~\cite{kerbl20233d}
    & 16.60 & 0.710 & 0.313
    & 21.71 & 0.808 & \cellcolor{orange!25}0.186
    & 24.55 & \cellcolor{orange!25}0.845 & \cellcolor{orange!25}0.155 \\
GuidedVD-3DGS~\cite{zhong2025taming}
    & \cellcolor{orange!25}19.93 & \cellcolor{orange!25}0.759 & \cellcolor{orange!25}0.297
    & \cellcolor{orange!25}22.98 & \cellcolor{orange!25}0.815 & 0.204
    & \cellcolor{orange!25}24.65 & 0.843 & 0.159 \\
Ours
    & \cellcolor{red!25}20.00 & \cellcolor{red!25}0.765 & \cellcolor{red!25}0.268
    & \cellcolor{red!25}23.41 & \cellcolor{red!25}0.835 & \cellcolor{red!25}0.181
    & \cellcolor{red!25}25.17 & \cellcolor{red!25}0.860 & \cellcolor{red!25}0.152 \\
\bottomrule
\end{tabular}
\end{table*}

\section{More Qualitative Comparison}
\label{sec:qual}

We provide additional qualitative results across 3-, 6-, and 9-view settings on Replica~\cite{straub2019replica} and ScanNet++~\cite{yeshwanth2023scannet++} datasets, as shown in Fig.~\ref{fig:comparison}. We compare against 3DGS~\cite{kerbl20233d}, FSGS~\cite{zhu2024fsgs}, InstantSplat~\cite{fan2024instantsplat}, DiFix3D~\cite{wu2025difix3d+}, GenFusion~\cite{wu2025genfusion}, and GuidedVD-3DGS~\cite{zhong2025taming} using the same baseline configurations as described in the main paper.

As illustrated in Fig~\ref{fig:comparison}, even with extremely sparse inputs (3 views), our method produces reasonable content and geometry while maintaining consistency. Compared to baselines, our approach demonstrates better scene coverage with fewer missing regions (black holes), improved geometric consistency with reduced ghosting artifacts, and overall higher visual quality. These improvements are particularly evident in challenging regions highlighted by white boxes.

\begin{figure*}[h!]
    \centering
    \includegraphics[width=\linewidth]{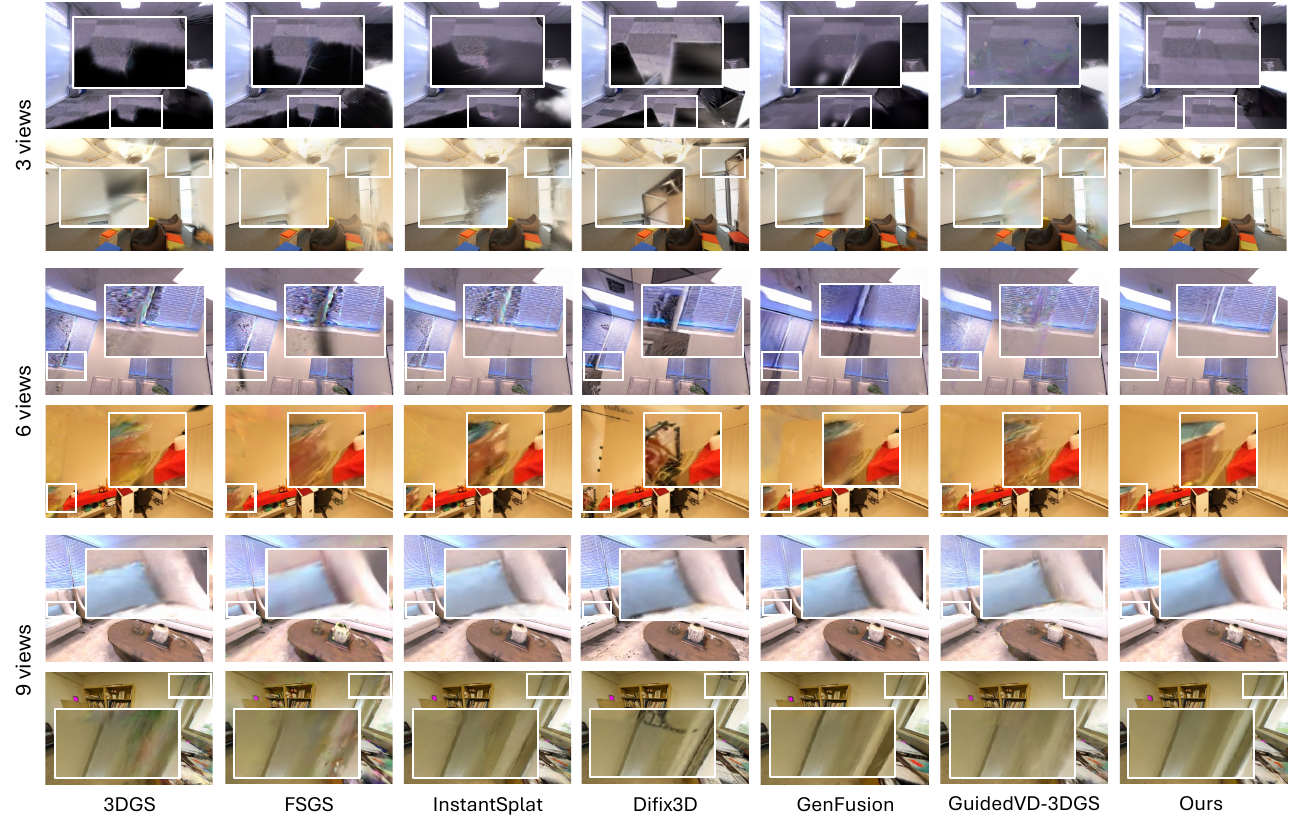}
    \vspace{-6mm}
    \caption{
        \textbf{Qualitative comparison on Replica~\cite{straub2019replica} and ScanNet++~\cite{yeshwanth2023scannet++} with 3, 6, and 9 sparse views.}
        Our method produces better coverage, geometric consistency, and fewer artifacts compared to baselines.
        White boxes highlight challenging regions. Best viewed zoomed in.
    }
    \label{fig:comparison}
\end{figure*}

\section{Runtime Analysis}
\label{sec:runtime}

\begin{wrapfigure}{r}{0.45\textwidth}
    \vspace{-25pt}
    \centering
    \footnotesize
    \setlength{\abovecaptionskip}{0pt}
    \setlength{\belowcaptionskip}{2pt}
    \captionof{table}{\textbf{Runtime breakdown of our method on (RTX 4090).}}
    \label{tab:runtime}
    \resizebox{\linewidth}{!}{
    \setlength{\tabcolsep}{4pt}
    \renewcommand{\arraystretch}{1.15}
    \begin{tabular}{lcc}
        \toprule
        Stage & Time (s) & Time (min) \\
        \midrule
        Coarse 3DGS init.\ \& render & 118 & 1.97 \\
        Multi-view outpainting & 93 & 1.55 \\
        3DGS refine (train + render) & 280 & 4.67 \\
        \midrule
        Total & 491 & 8.18 \\
        \bottomrule
    \end{tabular}
    }
    \vspace{4pt}
    \captionof{table}{\textbf{End-to-end runtime comparison (RTX A6000).}}
    \label{tab:runtime_a6000}
    \resizebox{\linewidth}{!}{
    \setlength{\tabcolsep}{4pt}
    \renewcommand{\arraystretch}{1.15}
    \begin{tabular}{lc}
        \toprule
        Method & Total Time \\
        \midrule
        GuidedVD-3DGS~\cite{zhong2025taming} & $\sim$3h 20min \\
        Ours & \textbf{9min 54s} \\
        \bottomrule
    \end{tabular}
    }
    \vspace{-30pt}
\end{wrapfigure}

We report the end-to-end runtime of our pipeline on a representative indoor scene (Replica\_6, \textit{office\_2}) with 6 input views at $512 \times 384$ resolution, evaluated on a single NVIDIA RTX 4090 GPU. The pipeline consists of three main stages: coarse 3DGS reconstruction and rendering, multi-view diffusion outpainting, and the final 3DGS refinement stage that incorporates DUSt3R point cloud initialization and refined 3DGS training.

Since GuidedVD-3DGS requires significantly larger GPU memory, it is evaluated on an NVIDIA RTX A6000 GPU. For a fair comparison, we additionally report the runtime of our method on the same hardware. On the Replica \textit{office\_2} scene, our pipeline takes \textbf{9 min 54 s}, while GuidedVD-3DGS requires approximately \textbf{3 hours 20 minutes}. This demonstrates that our method achieves substantially faster reconstruction while maintaining competitive reconstruction quality.

\section{Failure Cases}
\label{sec:failure}
\begin{figure}[t]
\centering
\includegraphics[width=\linewidth]{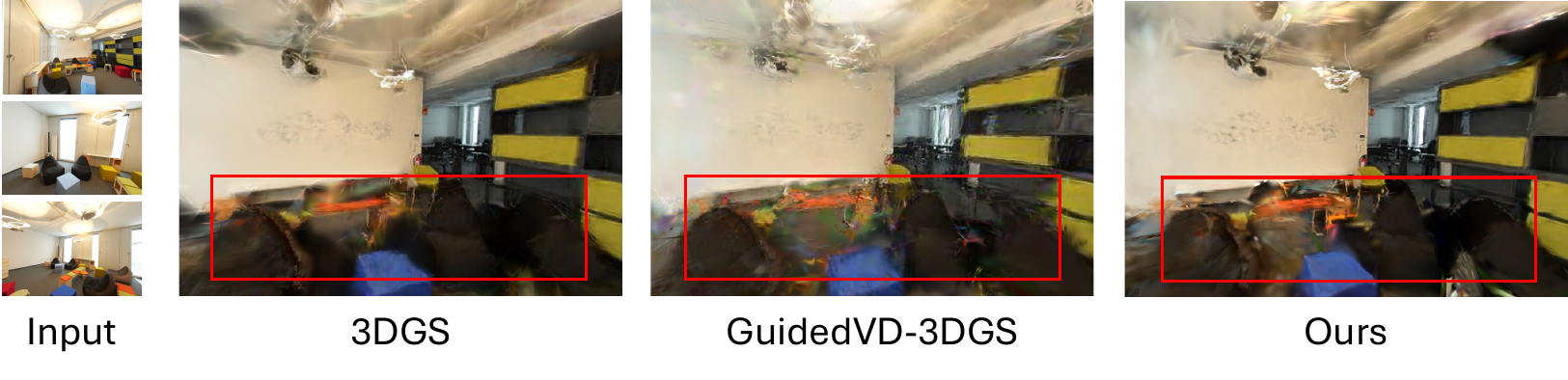}
    \vspace{-6mm}
\caption{
\textbf{Failure cases in heavily occluded regions.}
Due to severe occlusions in the scene, certain regions are never observed across all input views. Both outpainting (ours) and novel view generation 
methods struggle to reconstruct these completely unobserved areas. Red boxes highlight the occluded regions where reconstruction fails.
}
\label{fig:failure_cases}
\end{figure}

While our method demonstrates strong performance across a wide range of scenarios, it remains limited in scenes containing severe occlusions. This limitation is inherent to all methods that generate novel views for 3D reconstruction; current multi-view diffusion models face the same challenge, as even densely sampled novel viewpoints struggle to reconstruct regions that are severely occluded by obstacles. As shown in Fig.~\ref{fig:failure_cases}, such heavily occluded areas remain challenging for both geometry-aware outpainting and multi-view diffusion methods.

\noindent\textbf{Potential Solutions.}
A promising direction to address viewpoint-specific occlusions is to generate outpainted views from alternative camera perspectives with geometry-aware mechanisms, such as bird's-eye or top-down views with a larger FOV. By generating content from drastically different viewing angles, these views could potentially observe regions that are occluded from the original camera poses, thereby providing complementary supervision for the occluded areas.

\section{Baseline Implementation Details}
\label{sec:baseline}

In this section, we describe the implementation details of all baseline methods used in our experiments. Unless otherwise stated, we follow the official implementations and recommended settings provided by the authors. For most methods, the input camera poses are provided and the reconstruction is initialized using point clouds estimated by DUSt3R~\cite{wang2024dust3r}. InstantSplat follows its original pipeline and uses MASt3R~\cite{duisterhof2025mastrsfm} for initialization. Method-specific differences (e.g., initialization strategies) are described in the corresponding subsections.


\paragraph{3DGS~\cite{kerbl20233d}.}
We use the vanilla 3D Gaussian Splatting implementation~\cite{kerbl20233d} with its default training settings. 
The Gaussian primitives are initialized using the DUSt3R point cloud described above, and the input camera poses are kept fixed during optimization.

\paragraph{FSGS~\cite{zhu2024fsgs}.}
For FSGS, we follow the official implementation and training configuration provided by the authors~\cite{zhu2024fsgs}. In particular, we adopt the version that incorporates the depth supervision term, as suggested in the original implementation. All other training parameters remain consistent with the default configuration.

\paragraph{InstantSplat~\cite{fan2024instantsplat}.}
InstantSplat is implemented using the official codebase~\cite{fan2024instantsplat} with its default settings. The Gaussian initialization is obtained from point clouds reconstructed using MASt3R~\cite{duisterhof2025mastrsfm}, following the standard pipeline described in the original work.

\paragraph{Difix3D+~\cite{wu2025difix3d+}.}
The Difix3D framework contains multiple variants. In the original implementation, \textit{DiFix3D} and \textit{DiFix3D+} correspond to different variants within the same framework~\cite{wu2025difix3d+}, differing in whether a time neural enhancer is used during inference. In our experiments, we adopt the version without the time neural enhancer to maintain consistency with the base reconstruction pipeline.

\paragraph{GenFusion~\cite{wu2025genfusion}.}
GenFusion is evaluated using the official implementation and recommended hyperparameters~\cite{wu2025genfusion}. We follow the standard training and inference pipeline without modifying any of the default settings.

\paragraph{GuidedVD-3DGS~\cite{zhong2025taming}.}
GuidedVD-3DGS is implemented using the official codebase~\cite{zhong2025taming}. However, the results reported in the original paper use a different rendering resolution compared to our evaluation protocol. This difference leads to minor numerical discrepancies between the reported numbers in the paper and those reproduced in our environment. For transparency and fair comparison, we report both the original results from the paper and the results reproduced under our experimental settings.

\section{Per-Scene Quantitative Results}
\label{sec:per_scene}

We provide detailed per-scene quantitative results for every method across all datasets and view settings. Tab.~\ref{tab:scannet_replica_3views_per_scene} reports results on ScanNet++ and Replica with 3 input views, Tab.~\ref{tab:scannet_replica_6views_per_scene} with 6 input views, and Tab.~\ref{tab:mipnerf360_per_scene} reports results on Mip-NeRF360 with 6 and 9 input views. Each entry reports PSNR, SSIM, and LPIPS from top to bottom. Our method achieves consistently strong performance across scenes and datasets, ranking first or second in the majority of per-scene metrics while maintaining competitive results throughout, all while being \textbf{20$\times$ faster} than the strongest competing method.

\begin{table}[H]
\centering
\tiny
\setlength{\tabcolsep}{2pt}
\renewcommand{\arraystretch}{1.25}
\caption{\textbf{Per-scene quantitative results on ScanNet++ and Replica (3 views).}
Each entry reports PSNR$\uparrow$, SSIM$\uparrow$, and LPIPS$\downarrow$ from top to bottom.}
\vspace{-4pt}
\label{tab:scannet_replica_3views_per_scene}

\begin{tabular}{l!{\vrule}ccccc!{\vrule}ccccccc}
\toprule
& \multicolumn{5}{c!{\vrule}}{\textsc{ScanNet++ (3 Views)}}
& \multicolumn{7}{c}{\textsc{Replica (3 Views)}} \\
\midrule
Method
& 8a20 & 94ee & 7831 & a29c & avg
& office\_2 & office\_3 & office\_4 & room\_0 & room\_1 & room\_2 & avg \\
\midrule
3DGS
& \makecell[c]{15.63\\0.640\\0.340}
& \makecell[c]{16.61\\0.747\\0.283}
& \makecell[c]{19.08\\0.724\\0.311}
& \makecell[c]{14.68\\0.673\\0.345}
& \makecell[c]{16.50\\0.696\\0.320}
& \makecell[c]{18.87\\0.853\\0.147}
& \makecell[c]{22.51\\0.870\\0.104}
& \makecell[c]{20.96\\0.852\\0.131}
& \makecell[c]{20.96\\0.852\\0.131}
& \makecell[c]{22.14\\0.829\\0.129}
& \makecell[c]{19.34\\0.757\\0.223}
& \makecell[c]{20.39\\0.818\\0.154} \\
\midrule
FSGS
& \makecell[c]{18.09\\0.689\\0.299}
& \makecell[c]{15.97\\0.718\\0.355}
& \makecell[c]{17.44\\0.666\\0.412}
& \makecell[c]{14.82\\0.688\\0.369}
& \makecell[c]{16.58\\0.690\\0.359}
& \makecell[c]{21.07\\0.872\\0.177}
& \makecell[c]{22.31\\0.864\\0.117}
& \makecell[c]{20.24\\0.846\\0.140}
& \makecell[c]{18.40\\0.719\\0.213}
& \makecell[c]{21.81\\0.801\\0.167}
& \makecell[c]{21.20\\0.789\\0.220}
& \makecell[c]{20.84\\0.815\\0.172} \\
\midrule
InstantSplat
& \makecell[c]{17.27\\0.733\\0.232}
& \makecell[c]{17.42\\0.754\\0.311}
& \makecell[c]{17.79\\0.694\\0.399}
& \makecell[c]{14.38\\0.698\\0.317}
& \makecell[c]{16.72\\0.720\\0.315}
& \makecell[c]{19.78\\0.862\\0.124}
& \makecell[c]{22.97\\0.879\\0.096}
& \makecell[c]{22.43\\0.872\\0.103}
& \makecell[c]{18.05\\0.729\\0.177}
& \makecell[c]{22.62\\0.822\\0.130}
& \makecell[c]{18.05\\0.726\\0.219}
& \makecell[c]{20.65\\0.815\\0.142} \\
\midrule
Difix3D+
& \makecell[c]{16.05\\0.656\\0.297}
& \makecell[c]{16.05\\0.692\\0.364}
& \makecell[c]{16.38\\0.650\\0.355}
& \makecell[c]{14.09\\0.636\\0.366}
& \makecell[c]{15.64\\0.659\\0.346}
& \makecell[c]{19.58\\0.834\\0.203}
& \makecell[c]{20.77\\0.838\\0.145}
& \makecell[c]{20.65\\0.836\\0.138}
& \makecell[c]{17.89\\0.712\\0.232}
& \makecell[c]{18.79\\0.737\\0.201}
& \makecell[c]{19.14\\0.742\\0.245}
& \makecell[c]{19.47\\0.783\\0.194} \\
\midrule
GenFusion
& \makecell[c]{20.58\\0.747\\0.356}
& \makecell[c]{16.63\\0.725\\0.344}
& \makecell[c]{17.46\\0.693\\0.376}
& \makecell[c]{17.22\\0.734\\0.340}
& \makecell[c]{17.97\\0.725\\0.354}
& \makecell[c]{22.43\\0.882\\0.165}
& \makecell[c]{23.53\\0.871\\0.138}
& \makecell[c]{23.46\\0.861\\0.156}
& \makecell[c]{19.40\\0.745\\0.215}
& \makecell[c]{24.18\\0.835\\0.142}
& \makecell[c]{21.02\\0.803\\0.216}
& \makecell[c]{22.34\\0.833\\0.172} \\
\midrule
GuidedVD-3DGS
& \makecell[c]{17.32\\0.642\\0.342}
& \makecell[c]{\textbf{19.04}\\0.756\\0.316}
& \makecell[c]{19.23\\0.716\\0.359}
& \makecell[c]{19.70\\0.764\\0.306}
& \makecell[c]{18.82\\0.720\\0.328}
& \makecell[c]{25.07\\0.905\\0.114}
& \makecell[c]{\textbf{26.80}\\0.910\\0.097}
& \makecell[c]{\textbf{25.63}\\\textbf{0.898}\\0.109}
& \makecell[c]{\textbf{24.14}\\\textbf{0.791}\\0.171}
& \makecell[c]{\textbf{26.84}\\\textbf{0.854}\\0.126}
& \makecell[c]{\textbf{23.08}\\0.825\\0.208}
& \makecell[c]{\textbf{25.26}\\0.864\\0.138} \\
\midrule
Ours
& \makecell[c]{\textbf{20.77}\\\textbf{0.753}\\\textbf{0.225}}
& \makecell[c]{18.81\\\textbf{0.770}\\\textbf{0.273}}
& \makecell[c]{\textbf{20.35}\\\textbf{0.748}\\\textbf{0.299}}
& \makecell[c]{\textbf{20.31}\\\textbf{0.766}\\\textbf{0.263}}
& \makecell[c]{\textbf{20.06}\\\textbf{0.759}\\\textbf{0.265}}
& \makecell[c]{\textbf{25.13}\\\textbf{0.913}\\\textbf{0.088}}
& \makecell[c]{25.68\\\textbf{0.911}\\\textbf{0.075}}
& \makecell[c]{24.57\\0.897\\\textbf{0.095}}
& \makecell[c]{23.71\\0.789\\\textbf{0.150}}
& \makecell[c]{25.17\\0.851\\\textbf{0.114}}
& \makecell[c]{22.14\\\textbf{0.827}\\\textbf{0.179}}
& \makecell[c]{24.40\\\textbf{0.865}\\\textbf{0.117}} \\
\bottomrule
\end{tabular}
\end{table}

\begin{table}[t]
\centering
\tiny
\setlength{\tabcolsep}{2pt}
\renewcommand{\arraystretch}{1.25}

\caption{\textbf{Per-scene quantitative results on ScanNet++ and Replica (6views).}
Each entry reports PSNR$\uparrow$, SSIM$\uparrow$, and LPIPS$\downarrow$ from top to bottom.}
\vspace{-4pt}

\label{tab:scannet_replica_6views_per_scene}

\begin{tabular}{l!{\vrule}ccccc!{\vrule}ccccccc}
\toprule

& \multicolumn{5}{c!{\vrule}}{\textsc{ScanNet++ (6 Views)}} 
& \multicolumn{7}{c}{\textsc{Replica (6 Views)}} \\

\midrule

Method
& 8a20 
& 94ee 
& 7831 
& a29c 
& avg 
& office\_2 
& office\_3 
& office\_4 
& room\_0 
& room\_1 
& room\_2 
& avg \\

\midrule

3DGS
& \makecell[c]{24.33\\0.868\\0.109}
& \makecell[c]{20.22\\0.819\\0.200}
& \makecell[c]{21.41\\0.765\\0.262}
& \makecell[c]{20.87\\0.821\\0.169}
& \makecell[c]{21.71\\0.818\\0.186}
& \makecell[c]{27.40\\0.922\\0.070}
& \makecell[c]{25.05\\0.889\\0.101}
& \makecell[c]{23.66\\0.866\\0.135}
& \makecell[c]{21.92\\0.794\\0.142}
& \makecell[c]{24.03\\0.846\\0.140}
& \makecell[c]{24.38\\0.854\\0.156}
& \makecell[c]{24.74\\0.862\\0.124} \\

\midrule

FSGS
& \makecell[c]{23.56\\0.831\\0.234}
& \makecell[c]{20.72\\0.755\\0.402}
& \makecell[c]{20.86\\0.806\\0.309}
& \makecell[c]{21.60\\0.812\\0.246}
& \makecell[c]{21.69\\0.801\\0.298}
& \makecell[c]{25.99\\0.903\\0.092}
& \makecell[c]{24.24\\0.877\\0.114}
& \makecell[c]{23.51\\0.866\\0.148}
& \makecell[c]{21.65\\0.772\\0.171}
& \makecell[c]{22.89\\0.820\\0.171}
& \makecell[c]{23.18\\0.836\\0.175}
& \makecell[c]{23.91\\0.846\\0.145} \\

\midrule

InstantSplat
& \makecell[c]{22.84\\0.844\\0.121}
& \makecell[c]{20.27\\0.815\\0.209}
& \makecell[c]{21.16\\0.761\\0.275}
& \makecell[c]{20.50\\0.822\\0.165}
& \makecell[c]{21.19\\0.811\\0.193}
& \makecell[c]{26.62\\0.913\\0.071}
& \makecell[c]{20.04\\0.850\\0.187}
& \makecell[c]{22.72\\0.864\\0.150}
& \makecell[c]{21.35\\0.782\\0.138}
& \makecell[c]{23.65\\0.835\\0.142}
& \makecell[c]{24.14\\0.850\\0.150}
& \makecell[c]{23.09\\0.849\\0.140} \\

\midrule

Difix3D+
& \makecell[c]{23.86\\0.811\\0.175}
& \makecell[c]{19.42\\0.765\\0.269}
& \makecell[c]{18.86\\0.695\\0.312}
& \makecell[c]{20.34\\0.784\\0.219}
& \makecell[c]{20.62\\0.764\\0.244}
& \makecell[c]{23.40\\0.862\\0.160}
& \makecell[c]{21.70\\0.845\\0.149}
& \makecell[c]{22.79\\0.831\\0.176}
& \makecell[c]{20.42\\0.747\\0.204}
& \makecell[c]{21.22\\0.776\\0.219}
& \makecell[c]{21.62\\0.800\\0.218}
& \makecell[c]{21.86\\0.810\\0.188} \\

\midrule

GenFusion
& \makecell[c]{24.69\\0.860\\0.127}
& \makecell[c]{21.01\\0.818\\0.211}
& \makecell[c]{20.21\\0.750\\0.312}
& \makecell[c]{21.91\\0.805\\0.221}
& \makecell[c]{21.96\\0.808\\0.218}
& \makecell[c]{25.82\\0.909\\0.085}
& \makecell[c]{23.42\\0.869\\0.127}
& \makecell[c]{24.91\\0.878\\0.139}
& \makecell[c]{23.57\\0.808\\0.149}
& \makecell[c]{23.04\\0.830\\0.165}
& \makecell[c]{23.14\\0.837\\0.187}
& \makecell[c]{23.98\\0.855\\0.142} \\

\midrule

GuidedVD-3DGS$^\dagger$
& \makecell[c]{\textcolor{gray}{25.10}\\\textcolor{gray}{0.882}\\\textcolor{gray}{0.118}}
& \makecell[c]{\textcolor{gray}{23.10}\\\textcolor{gray}{0.860}\\\textcolor{gray}{0.201}}
& \makecell[c]{\textcolor{gray}{22.16}\\\textcolor{gray}{0.803}\\\textcolor{gray}{0.269}}
& \makecell[c]{\textcolor{gray}{25.21}\\\textcolor{gray}{0.857}\\\textcolor{gray}{0.157}}
& \makecell[c]{\textcolor{gray}{23.89}\\\textcolor{gray}{0.850}\\\textcolor{gray}{0.182}}
& \makecell[c]{\textcolor{gray}{27.46}\\\textcolor{gray}{0.916}\\\textcolor{gray}{0.083}}
& \makecell[c]{\textcolor{gray}{26.81}\\\textcolor{gray}{0.902}\\\textcolor{gray}{0.099}}
& \makecell[c]{\textcolor{gray}{27.43}\\\textcolor{gray}{0.897}\\\textcolor{gray}{0.122}}
& \makecell[c]{\textcolor{gray}{24.85}\\\textcolor{gray}{0.796}\\\textcolor{gray}{0.145}}
& \makecell[c]{\textcolor{gray}{26.00}\\\textcolor{gray}{0.851}\\\textcolor{gray}{0.142}}
& \makecell[c]{\textcolor{gray}{25.53}\\\textcolor{gray}{0.872}\\\textcolor{gray}{0.142}}
& \makecell[c]{\textcolor{gray}{26.35}\\\textcolor{gray}{0.872}\\\textcolor{gray}{0.122}} \\

\midrule

GuidedVD-3DGS$^\ddagger$
& \makecell[c]{24.18\\0.844\\0.143}
& \makecell[c]{22.11\\0.824\\0.219}
& \makecell[c]{\textbf{21.84}\\0.768\\0.283}
& \makecell[c]{23.79\\0.824\\0.172}
& \makecell[c]{22.98\\0.815\\0.204}
& \makecell[c]{\textbf{28.04}\\0.925\\0.087}
& \makecell[c]{\textbf{26.46}\\\textbf{0.901}\\0.112}
& \makecell[c]{\textbf{27.63}\\0.900\\0.130}
& \makecell[c]{25.55\\0.823\\0.143}
& \makecell[c]{\textbf{26.18}\\\textbf{0.858}\\0.157}
& \makecell[c]{\textbf{26.24}\\0.871\\0.167}
& \makecell[c]{\textbf{26.68}\\0.880\\0.133} \\

\midrule

Ours
& \makecell[c]{\textbf{24.98}\\\textbf{0.877}\\\textbf{0.104}}
& \makecell[c]{\textbf{22.38}\\\textbf{0.840}\\\textbf{0.186}}
& \makecell[c]{21.56\\\textbf{0.783}\\\textbf{0.269}}
& \makecell[c]{\textbf{24.70}\\\textbf{0.841}\\\textbf{0.164}}
& \makecell[c]{\textbf{23.41}\\\textbf{0.835}\\\textbf{0.181}}
& \makecell[c]{28.18\\\textbf{0.927}\\\textbf{0.062}}
& \makecell[c]{26.23\\0.900\\\textbf{0.087}}
& \makecell[c]{26.85\\\textbf{0.900}\\\textbf{0.104}}
& \makecell[c]{\textbf{25.65}\\\textbf{0.831}\\\textbf{0.117}}
& \makecell[c]{25.47\\0.857\\\textbf{0.127}}
& \makecell[c]{26.00\\\textbf{0.878}\\\textbf{0.128}}
& \makecell[c]{26.40\\\textbf{0.882}\\\textbf{0.104}} \\

\bottomrule
\end{tabular}

\vspace{-4pt}
\begin{flushleft}
\scriptsize $^\dagger$ Reported in paper, $^\ddagger$ Our reproduction.
\end{flushleft}

\end{table}

\begin{table}[H]
\centering
\tiny
\setlength{\tabcolsep}{1.4pt}
\renewcommand{\arraystretch}{1.1}
\caption{\textbf{Per-scene quantitative results on Mip-NeRF360 with 6 and 9 input views.}
Each entry reports PSNR$\uparrow$, SSIM$\uparrow$, and LPIPS$\downarrow$ from top to bottom.}
\vspace{-4pt}
\label{tab:mipnerf360_per_scene}

\resizebox{\textwidth}{!}{
\begin{tabular}{l!{\vrule}ccccccccc!{\vrule}c!{\vrule}ccccccccc!{\vrule}c}

\toprule
& \multicolumn{10}{c!{\vrule}}{\textsc{Mip-NeRF360 (6 Views)}} 
& \multicolumn{10}{c}{\textsc{Mip-NeRF360 (9 Views)}} \\
\midrule

Method
& bicycle & bonsai & counter & flowers & garden & kitchen & room & stump & treehill & avg
& bicycle & bonsai & counter & flowers & garden & kitchen & room & stump & treehill & avg \\

\midrule

3DGS
& \makecell{15.53\\0.239\\0.510}
& \makecell{13.69\\0.338\\0.551}
& \makecell{15.22\\0.420\\0.415}
& \makecell{13.14\\0.163\\0.539}
& \makecell{17.02\\0.403\\0.312}
& \makecell{17.51\\0.543\\0.290}
& \makecell{14.99\\0.479\\0.448}
& \makecell{16.07\\0.188\\0.511}
& \makecell{13.69\\0.308\\0.558}
& \makecell{15.21\\0.342\\0.459}

& \makecell{14.69\\0.263\\0.490}
& \makecell{14.89\\0.406\\0.461}
& \makecell{16.74\\0.504\\0.344}
& \makecell{13.95\\0.207\\0.472}
& \makecell{18.70\\0.477\\\textbf{0.241}}
& \makecell{18.30\\0.605\\0.274}
& \makecell{17.25\\0.585\\0.335}
& \makecell{17.75\\0.273\\0.420}
& \makecell{13.76\\0.344\\0.515}
& \makecell{16.23\\0.407\\0.395} \\

\midrule

FSGS
& \makecell{14.18\\0.193\\0.604}
& \makecell{13.39\\0.322\\0.588}
& \makecell{15.02\\0.409\\0.452}
& \makecell{12.49\\0.161\\0.653}
& \makecell{17.17\\0.397\\0.354}
& \makecell{17.14\\0.520\\0.355}
& \makecell{14.58\\0.472\\0.487}
& \makecell{15.75\\0.183\\0.556}
& \makecell{13.16\\0.278\\0.736}
& \makecell{14.76\\0.326\\0.532}

& \makecell{14.14\\0.251\\0.603}
& \makecell{14.36\\0.387\\0.508}
& \makecell{16.18\\0.486\\0.387}
& \makecell{13.60\\0.190\\0.636}
& \makecell{17.75\\0.419\\0.375}
& \makecell{15.02\\0.495\\0.448}
& \makecell{15.99\\0.554\\0.393}
& \makecell{17.37\\0.266\\0.463}
& \makecell{13.84\\0.307\\0.703}
& \makecell{15.36\\0.373\\0.502} \\

\midrule

InstantSplat
& \makecell{16.93\\\textbf{0.334}\\\textbf{0.457}}
& \makecell{13.89\\0.369\\0.550}
& \makecell{16.14\\0.463\\0.410}
& \makecell{\textbf{13.65}\\0.186\\0.546}
& \makecell{17.37\\0.405\\0.296}
& \makecell{18.26\\\textbf{0.606}\\\textbf{0.259}}
& \makecell{15.51\\0.511\\0.432}
& \makecell{\textbf{18.11}\\\textbf{0.277}\\0.459}
& \makecell{13.33\\0.342\\0.577}
& \makecell{15.91\\0.388\\0.443}

& \makecell{12.86\\0.291\\0.617}
& \makecell{14.84\\0.424\\0.482}
& \makecell{17.30\\0.526\\0.362}
& \makecell{13.38\\0.198\\0.571}
& \makecell{18.41\\0.446\\0.257}
& \makecell{\textbf{19.82}\\\textbf{0.666}\\\textbf{0.211}}
& \makecell{17.79\\0.601\\0.331}
& \makecell{\textbf{19.19}\\\textbf{0.358}\\0.405}
& \makecell{13.29\\0.385\\0.532}
& \makecell{16.32\\0.433\\0.419} \\

\midrule

Difix3D+
& \makecell{15.04\\0.187\\0.468}
& \makecell{13.47\\0.323\\\textbf{0.494}}
& \makecell{14.96\\0.397\\\textbf{0.380}}
& \makecell{12.95\\0.141\\\textbf{0.493}}
& \makecell{17.13\\0.374\\\textbf{0.291}}
& \makecell{17.37\\0.500\\0.281}
& \makecell{14.62\\0.450\\0.431}
& \makecell{15.39\\0.141\\\textbf{0.445}}
& \makecell{13.57\\0.263\\\textbf{0.491}}
& \makecell{14.94\\0.309\\\textbf{0.419}}

& \makecell{15.38\\0.219\\\textbf{0.451}}
& \makecell{14.55\\0.387\\\textbf{0.427}}
& \makecell{16.65\\0.484\\\textbf{0.323}}
& \makecell{13.83\\0.182\\\textbf{0.434}}
& \makecell{18.52\\0.437\\0.244}
& \makecell{17.98\\0.546\\0.259}
& \makecell{16.51\\0.536\\0.355}
& \makecell{16.71\\0.215\\\textbf{0.395}}
& \makecell{14.25\\0.308\\\textbf{0.451}}
& \makecell{16.04\\0.368\\\textbf{0.371}} \\

\midrule

GenFusion
& \makecell{16.29\\0.292\\0.570}
& \makecell{14.43\\0.389\\0.534}
& \makecell{\textbf{16.54}\\0.474\\0.420}
& \makecell{13.90\\\textbf{0.200}\\0.595}
& \makecell{18.23\\0.413\\0.379}
& \makecell{18.09\\0.521\\0.352}
& \makecell{17.24\\0.579\\0.407}
& \makecell{17.68\\0.226\\0.604}
& \makecell{\textbf{15.18}\\\textbf{0.363}\\0.613}
& \makecell{16.40\\0.384\\0.487}

& \makecell{\textbf{16.53}\\0.309\\0.537}
& \makecell{\textbf{15.29}\\\textbf{0.429}\\0.448}
& \makecell{\textbf{17.76}\\\textbf{0.547}\\0.364}
& \makecell{\textbf{14.91}\\\textbf{0.234}\\0.496}
& \makecell{19.38\\0.475\\0.302}
& \makecell{19.88\\0.586\\0.254}
& \makecell{19.03\\0.651\\0.307}
& \makecell{18.53\\0.290\\0.486}
& \makecell{\textbf{16.62}\\\textbf{0.393}\\0.488}
& \makecell{17.55\\0.435\\0.409} \\

\midrule

GuidedVD-3DGS
& \makecell{14.80\\0.189\\0.670}
& \makecell{12.75\\0.293\\0.722}
& \makecell{12.05\\0.305\\0.714}
& \makecell{12.48\\0.135\\0.717}
& \makecell{15.20\\0.210\\0.541}
& \makecell{14.32\\0.359\\0.658}
& \makecell{13.57\\0.462\\0.566}
& \makecell{15.76\\0.173\\0.602}
& \makecell{14.12\\0.333\\0.569}
& \makecell{13.89\\0.273\\0.640}

& \makecell{14.49\\0.238\\0.519}
& \makecell{14.56\\0.383\\0.483}
& \makecell{16.81\\0.506\\0.346}
& \makecell{13.49\\0.194\\0.492}
& \makecell{18.52\\0.456\\0.253}
& \makecell{15.81\\0.542\\0.373}
& \makecell{17.57\\0.598\\0.327}
& \makecell{17.27\\0.246\\0.432}
& \makecell{13.44\\0.313\\0.538}
& \makecell{15.77\\0.386\\0.418} \\

\midrule

Ours
& \makecell{\textbf{17.33}\\0.285\\0.476}
& \makecell{\textbf{14.74}\\\textbf{0.392}\\0.524}
& \makecell{16.45\\\textbf{0.481}\\0.398}
& \makecell{13.55\\0.192\\0.547}
& \makecell{\textbf{18.84}\\\textbf{0.430}\\0.294}
& \makecell{\textbf{18.68}\\0.567\\0.304}
& \makecell{\textbf{17.93}\\\textbf{0.615}\\\textbf{0.370}}
& \makecell{17.72\\0.231\\0.488}
& \makecell{15.39\\0.341\\0.522}
& \makecell{\textbf{16.74}\\\textbf{0.393}\\0.436}

& \makecell{16.21\\\textbf{0.314}\\\textbf{0.450}}
& \makecell{16.15\\0.461\\0.444}
& \makecell{17.34\\0.539\\0.365}
& \makecell{14.98\\0.229\\0.463}
& \makecell{\textbf{19.89}\\\textbf{0.492}\\0.244}
& \makecell{19.38\\0.626\\0.266}
& \makecell{\textbf{20.55}\\\textbf{0.700}\\\textbf{0.282}}
& \makecell{18.66\\0.289\\0.423}
& \makecell{14.87\\0.384\\0.488}
& \makecell{\textbf{17.56}\\\textbf{0.448}\\0.381} \\

\bottomrule
\end{tabular}
}
\end{table}

\end{document}